\documentclass[10pt,twocolumn,letterpaper]{article}

\usepackage{iccv}
\usepackage{times}
\usepackage{epsfig}
\usepackage{graphicx}
\usepackage{amsmath}
\usepackage{amssymb}

\usepackage{authblk}  
\usepackage{makecell}
\usepackage{float}
\usepackage{multirow}
\usepackage{bm}
\usepackage{booktabs}
\usepackage{siunitx}
\usepackage{textcomp} 
\usepackage{pifont}
\usepackage[utf8]{inputenc}

\usepackage[pagebackref=true,breaklinks=true,letterpaper=true,colorlinks,bookmarks=false]{hyperref}

\iccvfinalcopy % *** Uncomment this line for the final submission

\def\iccvPaperID{7627} % *** Enter the ICCV Paper ID here

\ificcvfinal\fi

\begin{document}
	
	%%%%%%%%% TITLE
	\title{MSR-GCN: Multi-Scale Residual Graph Convolution Networks for Human Motion Prediction}
	
	\author[1]{Lingwei Dang}
	\author[1\thanks{Corresponding author: nieyongwei@scut.edu.cn}]{Yongwei Nie}
	\author[2]{Chengjiang Long}
	\author[3]{Qing Zhang}
	\author[1]{Guiqing Li}
	\affil[1]{School of Computer Science and Engineering, South 
		China University of Technology, China}
	\affil[2]{JD Finance America Corporation, USA}
	\affil[3]{School of Computer Science and Engineering, Sun Yat-sen University, China}
	
	\renewcommand\Authands{ and }
	
	\maketitle
	\ificcvfinal\thispagestyle{empty}\fi
	
	%%%%%%%%% ABSTRACT
	\begin{abstract}
		Human motion prediction is a challenging task due to the stochasticity and aperiodicity of future poses. Recently, graph convolutional network has been proven to be very effective to learn dynamic relations among pose joints, which is helpful for pose prediction. On the other hand, one can abstract a human pose recursively to obtain a set of poses at multiple scales. With the increase of the abstraction level, the motion of the pose becomes more stable, which benefits pose prediction too. In this paper, we propose a novel Multi-Scale Residual Graph Convolution Network (MSR-GCN) for human pose prediction task in the manner of end-to-end. The GCNs are used to extract features from fine to coarse scale and then from coarse to fine scale. The extracted features at each scale are then combined and decoded to obtain the residuals between the input and target poses. Intermediate supervisions are imposed on all the predicted poses, which enforces the network to learn more representative features. Our proposed approach is evaluated on two standard benchmark datasets, \ie, the Human3.6M dataset and the CMU Mocap dataset. Experimental results demonstrate that our method outperforms the state-of-the-art approaches. Code and pre-trained models are available at \href{https://github.com/Droliven/MSRGCN}{https://github.com/Droliven/MSRGCN}.
		
	\end{abstract}
	
	%%%%%%%%% BODY TEXT
	\section{Introduction}\label{sec:intro}
	
	\begin{figure}[t]
		\centering
		\includegraphics[width=0.8\linewidth ]{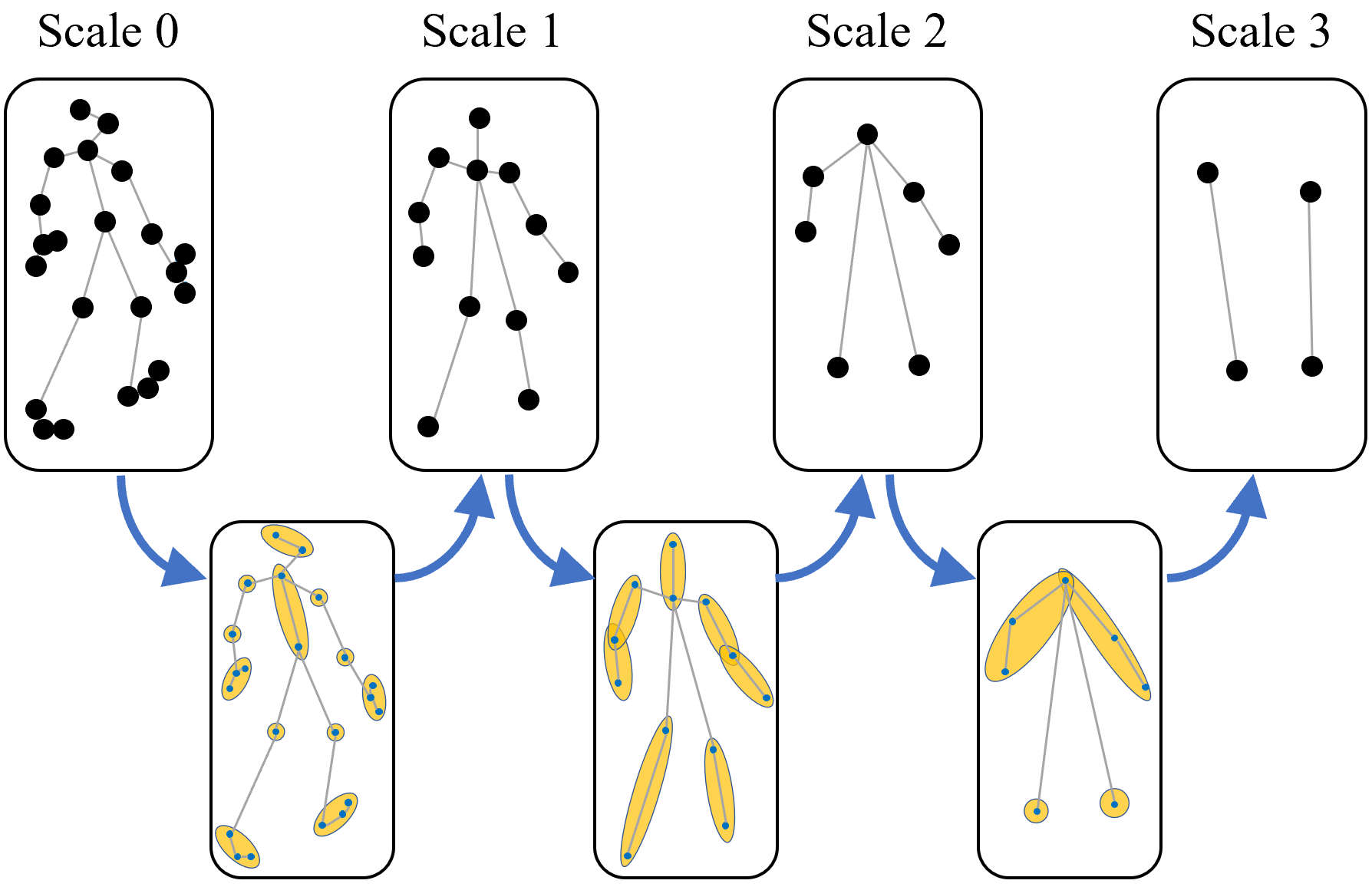}
		\caption{A human pose can be abstracted step by step to obtain a series of poses from fine to coarse scale, by grouping joints in close proximity together and replacing the group with a single joint.}
		\label{fig:fine-coarse}
		\vspace{-0.5cm}
	\end{figure}
	
	%------------------------------------------------------------------------
	Human motion prediction plays a critical role in many fields, such as human-computer interaction, autonomous driving, and video completion. Simple periodic motion patterns can be tackled by traditional methods such as hidden Markov model \cite{brand2000style}, linear dynamic system \cite{pavlovic2000learning}, restricted Boltzmann machine \cite{taylor2007modeling}, Gaussian process latent variable models \cite{wang2005gaussian} and random forests \cite{lehrmann2014efficient}, while more complex motion is intractable for these methods. The latest approaches are almost all data-driven methods with deep learning. However, considering the stochasticity and aperiodicity of human motion, it still remains a challenging task to predict accurate future motion in long term giving observed arbitrary poses. The main difficulty is how to model the spatiotemporal dependencies of human poses.
	
	Lots of prior efforts with Convolutional Neural Networks (CNNs)~\cite{yang2020sta,liu2020trajectorycnn}, Recurrent Neural Networks (RNNs) ~\cite{fragkiadaki2015recurrent,martinez2017human,song2017end,tang2018long,sang2020human,guo2019human,chan2020gas,al2020attention}, and Generative Adversarial Networks (GANs)
	~\cite{zheng2020towards,gui2018adversarial,ke2019learning,hernandez2019human,cui2021efficient,wang2019early,kundu2019bihmp}, have been made for tackling the challenging task. However, they neglect the inner-frame kinematic dependencies between body joints. Although they have achieved success in some cases, the prediction accuracy depends on the size of convolution filters and the stability of the frame-by-frame prediction.
	%In the past recent years,
	Nowadays, Graph Convolution Networks (GCNs) have been widely used in various fields as well as in the task of human motion prediction \cite{mao2019learning,li2020dynamic,cui2020learning,li2020multi,liu2020disentangling,zhang2020context,Shi_2019_CVPR}, which work very well for non-grid graph-structured data especially for skeleton-based 3D human pose sequences. %In particular, 
	Recently, Mao \etal~\cite{mao2019learning} jointly model spatial structure by GCNs with learnable connectivity and temporal information via discrete cosine transformation (DCT) to predict human motion.
	Li \etal~\cite{li2020dynamic} propose a dynamic multi-scale graph neural network within an encoder-decoder framework to extract deep features at multiple scales. 
	Although these two works exhibit promising results on benchmark datasets, there is still space to be explored for more high-quality human motion prediction.
	
	In this paper, we propose a Multi-Scale Residual Graph Convolution Network (MSR-GCN), as illustrated in Figure~\ref{fig:msgcn}, for 3D human motion prediction. By treating a human pose as a fully connected graph whose vertices are the pose joints, we employ graph convolution networks to dynamically learn the relations between all pairs of joints flexibly regardless of the physical distance between them. But GCN alone cannot capture the hierarchical structure of human pose~\cite{mao2019learning}. That is, as shown in Figure~\ref{fig:fine-coarse}, one can abstract a human pose by grouping joints in close proximity together and representing the group by just one joint, yielding a coarser pose. Since a group of joints usually come from the same body part, gradually abstracting body parts in this way can significantly stabilize the motion pattern of the body. We find that the motion in the coarser level is more stable for which the pose prediction is easier. It is promising to predict the poses in the coarsest level firstly, and then go up to finer levels gradually.
	
	Based on the above analysis, we compensate GCN with the capacity of modeling hierarchical and contextual information of human pose by designing multiple GCNs with a multi-scale architecture. A group of the GCNs forms a descending path to extract features from fine to coarse scale, followed by another group of GCNs that extract multi-scale features inversely along an ascending path. Based on these features, we predict poses at all scales and impose intermediate supervision for more representative features. 
	We also add residual connections between the input and the output poses as suggested by~\cite{mao2019learning}, making the whole framework learn residuals instead of the target poses directly.
	
	Note that Li \etal~\cite{li2020dynamic} have also observed this natural hierarchical structure of human pose, but they aim to extract rich features with the help of the multi-scale joint abstraction and then decode the future poses from the multi-scale features with a recurrent decoder. In contrast, the encoder and decoder in our method are organized in a U-Net-like multi-scale manner equipped with intermediate losses, differing from the multiscale strategy in ~\cite{li2020dynamic}.
	
	In short, our main technical contributions are as follows:
	\begin{itemize}
		\item We propose a novel multi-scale residual graph convolution network for human pose prediction in an end-to-end manner, which consists of multiple GCNs organized in a multi-scale architecture. 
		\item The well-designed descending and ascending GCN blocks can extract features in both fine-to-coarse and coarse-to-fine manners.
		\item The intermediate supervision imposed at each scale enforces to learn more representative features, benefiting high-quality future prediction.
	\end{itemize}
	
	%------------------------------------------------------------------------
	\section{Related work}
	
	{\bf Human motion prediction}. Many deep learning based methods have been proposed to handle human motion prediction. Existing CNN-based works like~\cite{yang2020sta,liu2020trajectorycnn} treat a pose sequence as a two-dimensional matrix where one axis is the spatial axis and another one indicates the temporal axis, then spatiotemporal convolutional filters can be used to the pose data like what has been done for an image. However, pose data, in essence, is very different from images, lacking repeated elements that give a high response to the same filter, thus reducing the effectiveness of the convolutions. 
	Although RNN-based methods like~\cite{fragkiadaki2015recurrent,martinez2017human,song2017end,tang2018long,sang2020human,guo2019human,chan2020gas,al2020attention} have advantages in dealing with time-related tasks, the discontinuity and error accumulation problems often happen because of the frame-by-frame prediction manner. Also, the training of RNN models is easy to collapse with gradient explosion or disappearing. What's more, these networks neglect the inner-frame kinematic dependencies between body joints. 
	Generative adversarial networks \cite{zheng2020towards,gui2018adversarial,ke2019learning,hernandez2019human,cui2021efficient,wang2019early,kundu2019bihmp} are deemed to generate realistic data whose pattern is similar to the training data. Nevertheless, they are vulnerable and require skillful training.
	Transformer-based networks like~\cite{cai2020learning, aksan2020spatio} are supposed to be capable of capturing long-range temporal dependencies directly but usually have quite high computing costs.
	
	{\bf Graph Convolution Networks (GCNs)} are suitable for tasks with non-grid and graph-structural data, \eg, biological gene, point cloud, human social network \cite{yan2018spatial}, and human motion prediction for the graph-structure nature of the human skeleton. 
	They have been successfully applied to many applications like visual recognition~\cite{hua2013collaborative, hong2020graph, Hu:TIP2021, Hu:arXiv2021, Long:ICCV2015, Long:IJCV2016, Long:CVPR2017, Hua:TPAMI2018}, object detection~\cite{wang2020joint, Islam:CVPR2020}, action localization~\cite{zeng2019graph, Islam:AAAI2021}, trajectory prediction~\cite{Shi:CVPR2021}, and image captioning~\cite{Dong:MM2021}. 
	In particular, since graph convolution is more inclined to capture spatial information, Si \etal~\cite{si2019attention} combines it with LSTM to enhance its capability of modeling temporal dependencies between human skeleton joints. Works of \cite{mao2019learning,li2019actional,cui2020learning} allow graph convolution network to learn relations between any pair of human joints. Mao \etal~\cite{mao2019learning} design a fully connected GCN to adaptively learn the necessary connectivity for the motion prediction task and apply discrete cosine transformation (DCT) to handle temporal information. Cui \etal~\cite{cui2020learning} enhance the role of natural connectivity of human joints among all the edges of the fully connected graph. Li \etal~\cite{li2020dynamic} propose a graph neural network with a multi-scale graph computational unit where features are extracted at a single individual scale and then fused across scales.
	Differently, we use GCNs at different scales to extract features for these scales separately. 
	
	%------------------------------------------------------------------------
	
	\section{Methodology}
	
\begin{figure*}
		\vspace{-0.5cm}
		\centering
		\includegraphics[width=0.9\linewidth]{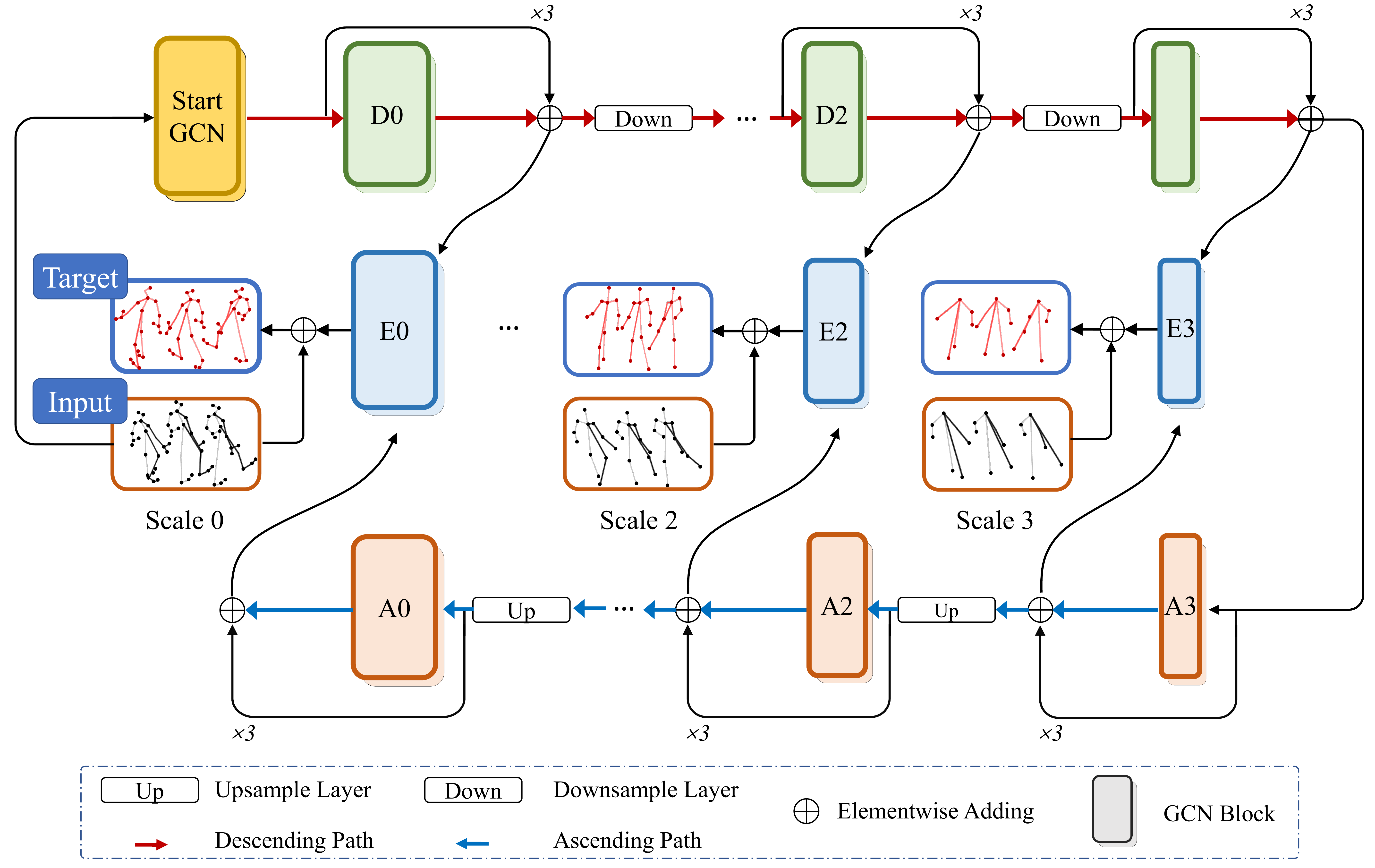}
		\caption{The architecture of the proposed MSR-GCN comprising one start GCN, four descending GCNs ($D0,D1,D2,D3$), four ascending GCNs ($A0,A1,A2,A3$), and four end GCNs ($E0,E1,E2,E3$). The start GCN takes the black poses at scale 0 as input. Then descending and ascending GCNs are stacked sequentially to extract features for each scale twice. The combined features at each scale are finally fed into the corresponding end GCN for decoding. Residual connections are added after every end GCN that add the ground truth poses to the output of each GCN, making the network learn residuals rather than the target poses directly.}
		\label{fig:msgcn}
		\vspace{-0.4cm}
	\end{figure*}

	Human pose prediction is a task to produce future pose sequence given the currently observed frames. Supposing the historical poses are $X_{1:T_{h}} = [X_{1},...,X_{T_{h}}] \in{\mathbb{R}^{J\times D \times T_{h}}}$ with $T_{h}$ frames, among which $X_{t}$ depicts a single 3D human pose with $J$ joints in the $D$-dimensional space (here $D$ is 3) at time $t$. Similarly, the future pose sequence with $T_{f}$ frames is defined as $X_{T_{h}+1:T_{h}+T_{f}}$. We need a model $\mathcal{F}_{predict}(\cdot)$ to predict the future unknown pose sequence $\hat{X}_{T_{h}+1:T_{h}+T_{f}}$ giving $X_{1:T_{h}}$ that approximates the ground truth $X_{T_{h}+1:T_{h}+T_{f}}$ as much as possible. We fulfill this task by proposing a novel Multi-Scale Residual Graph Convolution Network called MSR-GCN, as illustrated in Figure~\ref{fig:msgcn}. 
	
	In the following, the basic GCN model for pose prediction is introduced firstly, then the multi-scale architecture used to obtain superior prediction accuracy is shown.

	%------------------------------------------------------------------------
	\subsection{Basic GCNs}
	Firstly, we reformulate our prediction objective by rearranging the input and output pose sequences. Instead of performing prediction based on $X_{1:T_{h}}$, we replicate the last pose $X_{T_h}$ for $T_f$ times, obtaining a sequence of length $T=T_h+T_f$. We then use this sequence as the input to predict the future pose sequence comprising of $\hat{X}_{1:T_{h}}$ and $\hat{X}_{T_{h}+1:T_{h}+T_{f}}$. According to ~\cite{mao2019learning}, this prediction task can be translated to compute a residual vector between $\hat{X}_{1:T}$ and the ground truth ${X}_{1:T}$, which we also find very effective to improve the prediction accuracy.
	
	For pose prediction, it has been proven very useful to model the spatial structure of the poses~\cite{mao2019learning,cui2020learning}. This is because the spatial dependencies between human joints exhibit inherent and consistent characteristics over the whole action period, which is of great importance for human pose prediction. The dependencies that can be utilized are not confined to joints with kinematic links such as between elbow and wrist, but any pair of joints can affect each other. For example, when a person walks, the hands vibrate periodically, so it is essential to explore the dependencies of two hands for their predictions. GCN~\cite{kipf2016semi} is good at discovering these relationships by viewing a pose as a fully-connected graph with $K$ nodes, where $K=J\times D$, and an adjacency matrix $\textbf{A}\in{\mathbb{R}^{K\times K}}$ which represents the strength of edges of the graph is learned by the GCN.
	
	A GCN is usually composed of a set of graph convolutional layers that are sequentially stacked together. Formally, let $\textbf{H}^l\in{\mathbb{R}^{K\times F^l}}$ be the input to a graph convolutional layer, $\textbf{A}^l\in{\mathbb{R}^{K\times K}}$ the adjacency matrix, and $\textbf{W}^l\in{\mathbb{R}^{F^l\times F^{l+1}}}$ the trainable parameters, the output of the graph convolutional layer is:
	\begin{equation}
		\textbf{H}^{l+1} = \sigma(\textbf{A}^l\textbf{H}^l\textbf{W}^l),
	\end{equation}
	where $\textbf{H}^{l+1}\in{\mathbb{R}^{K\times F^{l+1}}}$, and $\sigma(\cdot)$ is an activation function.
	
	\begin{figure}[t]
		\vspace{-0.2cm}
		\centering
		\includegraphics[width=1.0\linewidth]{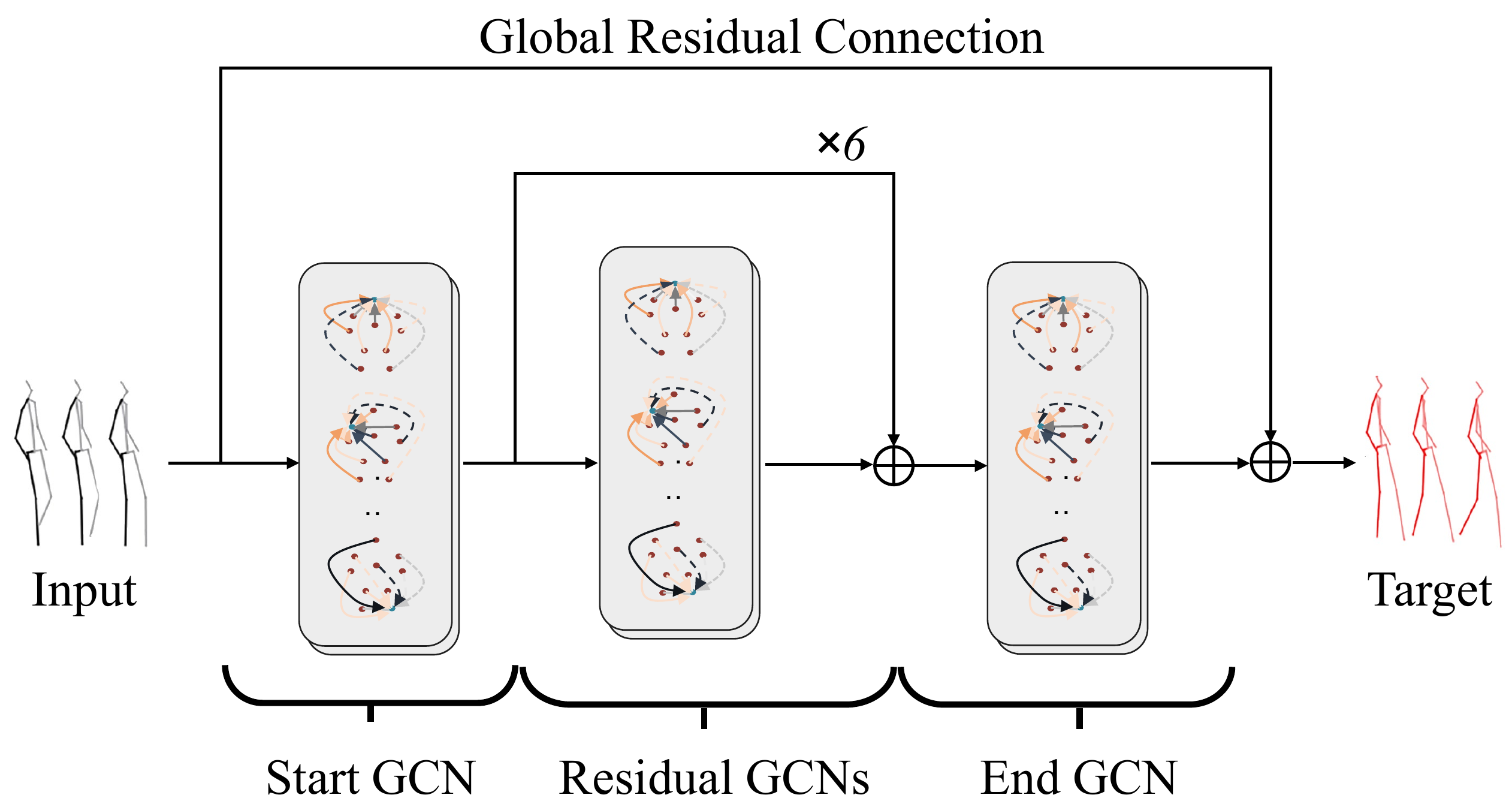}
		\caption{The basic GCN model for pose prediction comprising a start GCN, 6 residual GCNs, and an end GCN. The start GCN maps the input from pose space to feature space, the residual GCNs are used to extract features in the feature space, and finally, the end GCN maps the features back to the poses. A residual connection is added between the input and output poses, making the whole network learn residuals rather than the target poses directly.}
		\label{fig:gcn}
		\vspace{-0.2cm}
	\end{figure}
	
	To map the input pose sequence to the target pose sequence, we design one start GCN, one end GCN, and 6 residual GCNs, the architecture of which is shown in Figure~\ref{fig:gcn}. The start GCN has 2 graph convolutional layers, projecting the input pose sequence from the space of $\mathbb{R}^{K\times T}$ to $\mathbb{R}^{K\times F}$, with $F=256$ in this paper. Following are 6 residual GCNs each containing 2 graph convolutional layers which accept features in space $\mathbb{R}^{K\times F}$ and also output features in the same space. Finally, the end GCN, also containing 2 graph convolutional layers, projects the features in space $\mathbb{R}^{K\times F}$ to the target pose sequence in space $\mathbb{R}^{K\times T}$. The whole network learns the residual vector between the input and target pose sequences by adding a global skip connection as shown in Figure~\ref{fig:gcn}.
	
	Note that the above pose prediction network with basic GCNs is similar to the method proposed in ~\cite{mao2019learning} except for the Discrete Cosine Transform (DCT) and inverse DCT for data representation transformation. In this paper, we abandon the DCT transformations since directly computing global residuals between padded input poses and the target poses without translating to DCT coefficients is effective enough and computationally more efficient. In the following, we show how the basic architecture in Figure~\ref{fig:gcn} can be further improved by taking advantage of the multi-scale properties of human pose~\cite{li2020dynamic}.
	
	\subsection{Multi-scale Residual GCNs}
	
	Intuitively, a human pose can be simplified step by step to obtain a set of fine-to-coarse poses. With the increase of the coarse-scale, the motion of the pose becomes more stable, which usually means the pose prediction in this scale is easier than a finer scale. This motivates us to propose a Multi-scale Residual Graph Convolution Network (MSR-GCN), in which we perform prediction at the coarsest level firstly, and then go up to higher levels step by step. As shown in Figure~\ref{fig:msgcn}, our MSR-GCN is composed of four kinds of GCNs: one start GCN, a set of descending and ascending GCN blocks, and a set of end or decoding GCNs. 
	
	Before introducing MSR-GCN, let us describe how we abstract a human pose. As shown in the leftmost picture of Figure~\ref{fig:fine-coarse}, the finest pose has 22 joints. We abstract the finest pose recursively to obtain 3 poses with 12, 7, and 4 joints respectively. The subplots in the second row of Figure~\ref{fig:fine-coarse} (from left to right) depict how to combine the joints at the finer level, while those in the first row show the obtained poses at the next levels correspondingly. 
	Note that we also tried other grouping manners, but found this scheme yields the most stable motion at the coarsest level (see comparisons in Section ~\ref{sec:ablationgroupingmanner}).

	\textbf{Start GCN} is composed of 2 convolutional layers, mapping the input poses into the feature space. The pose space is $\mathbb{R}^{K\times T}$ as defined above, and the feature space is $\mathbb{R}^{K\times F}$ with $F=256$. We use the finest-scale pose sequence as the input to the start GCN while the pose sequences at other scales are only used at end GCNs to calculate residuals. 
	
	\textbf{Descending and ascending GCN blocks.} Since we have abstracted the human pose in four levels, we use four descending and four ascending GCN blocks, namely $D0,D1,D2,D3$ and $A3,A2,A1,A0$, to extract features at the four scales. Each of these blocks loops a residual GCN 6 times, and each GCN has 2 graph convolutional layers. The eight GCN blocks are sequentially stacked together. Along the whole descending and ascending path, the feature dimension  $F$ is always kept as 256, but the pose dimension $K$ changes between adjacent descending or ascending blocks. For example, $D0$ extracts features in space $\mathbb{R}^{K_0\times F}$ with $K_0 = 22\times 3=66$, while $K_1=36$, $K_2=21$ and $K_3=12$ for $D1$, $D2$ and $D3$. We use a downsampling layer to transform the features outputted by $D0$ into the space of $\mathbb{R}^{K_1\times F}$. The descending blocks gradually reduce the pose dimension which is then gradually increased by the ascending blocks with upsampling layers. We concatenate the features extracted by a descending GCN block and the corresponding ascending GCN block together and deliver them to the end GCNs for decoding.
	
	\textbf{End GCNs} are used for decoding the concatenated features extracted by descending and ascending blocks to poses. Like start GCN, an end GCN is also composed of 2 graph convolutional layers. But instead of just one start GCN, we design 4 end GCNs, namely $E0,E1,E2,E3$, to decode combined features at four different scales, respectively. 
	Intermediate supervisions by computing the L2 distances between the decoded poses and their ground truth at all scales are used to train the whole network, which is a commonly adopted strategy in many works~\cite{wei2016convolutional,zhang2020multi}. Ablation experiments show that with the intermediate supervisions, better prediction accuracy can be obtained, which we conjecture is due to the reason that it helps extract more representative features in coarser levels and enforce the whole network to learn the prediction from coarse to fine scale. The output of ``E0" is the predicted target pose sequence.
	
	\textbf{Residual Connections.} Besides the residual connections in descending and ascending GCNs, we add a residual connection after each end GCN. That is to say, we add the input pose sequence (at different scales) to the output of the end GCN. In this way, the MSR-GCN learns the residual vector between the input and ground truth at all levels.
	
	\subsection{Implementation Details}
	
	We choose Adam as the optimizer with the initial learning rate of 2e-4, which decays by 0.98 every two epochs and train the network on an NVIDIA RTX 3090 GPU card.

	\section{Experiments}
	To verify the effectiveness of MSR-GCN, we run experiments on two standard benchmark motion capture datasets, including Human3.6M (H3.6M) and CMU Mocap dataset. Here we first introduce the two datasets, the evaluation metric and the baselines we compare with, then present experimental results and ablation analysis.
	
	% human3.6m short-term
	\begin{table*}[]
		\begin{center}
			\renewcommand\tabcolsep{4.5pt}
			\scriptsize
			\caption{Comparisons for short-term prediction on 15 action categories of H3.6M and the averages. The best results are highlighted in bold.}
			\label{tab:h36m_short1}
			\begin{tabular}{c|cccc|cccc|cccc|cccc}
				\hline
				scenarios             & \multicolumn{4}{c|}{walking}                                      & \multicolumn{4}{c|}{eating}                                      & \multicolumn{4}{c|}{smoking}                                     & \multicolumn{4}{c}{discussion}                                   \\ \hline
				millisecond (ms)      & 80             & 160            & 320            & 400            & 80            & 160            & 320            & 400            & 80            & 160            & 320            & 400            & 80             & 160            & 320            & 400            \\ \hline
				Residual sup.~\cite{martinez2017human} & 29.36          & 50.82          & \ \ 76.03          & \ \ 81.51          & 16.84         & \ \ 30.60          & \ \ 56.92          & \ \ 68.65          & 22.96         & 42.64          & \ \ 70.14          & \ \ 82.68          & 32.94          & 61.18          & \ \ 90.92          & \ \ 96.19          \\
				DMGNN~\cite{li2020dynamic}        & 17.32          & 30.67          & \ \ 54.56          & \ \ 65.20          & 10.96         & \ \ 21.39          & \ \ 36.18          & \ \ 43.88          & \ \ 8.97          & 17.62          & \ \ 32.05          & \ \ 40.30          & 17.33          & 34.78          & \ \ 61.03          & \ \ 69.80          \\
				Traj-GCN~\cite{mao2019learning}     & 12.29          & 23.03          & \ \ 39.77          & \ \ 46.12          & \textbf{\ \ 8.36} & \textbf{\ \ 16.90} & \ \ 33.19          & \ \ 40.70          & \textbf{\ \ 7.94} & \textbf{16.24} & \ \ 31.90          & \ \ 38.90          & 12.50          & 27.40          & \ \ 58.51          & \ \ 71.68          \\
				MSR-GCN               & \textbf{12.16} & \textbf{22.65} & \textbf{\ \ 38.64} & \textbf{\ \ 45.24} & \ \ 8.39          & \ \ 17.05          & \textbf{\ \ 33.03} & \textbf{\ \ 40.43} & \ \ 8.02          & 16.27          & \textbf{\ \ 31.32} & \textbf{\ \ 38.15} & \textbf{11.98} & \textbf{26.76} & \textbf{\ \ 57.08} & \textbf{\ \ 69.74} \\ \hline
			\end{tabular}
			\begin{tabular}{c|cccc|cccc|cccc|cccc}
				\hline
				scenarios             & \multicolumn{4}{c|}{directions}                                  & \multicolumn{4}{c|}{greeting}                                     & \multicolumn{4}{c|}{phoning}                                      & \multicolumn{4}{c}{posing}                                       \\ \hline
				millisecond (ms)      & 80            & 160            & 320            & 400            & 80             & 160            & 320            & 400            & 80             & 160            & 320            & 400            & 80             & 160            & 320            & 400            \\ \hline
				Residual sup.~\cite{martinez2017human} & 35.36         & 57.27          & \ \ 76.30          & \ \ 87.67          & 34.46          & \ \ 63.36          & 124.60         & 142.50         & 37.96          & 69.32          & 115.00         & 126.73         & 36.10          & 69.12          & 130.46         & 157.08         \\
				DMGNN ~\cite{li2020dynamic}        & 13.14         & 24.62          & \ \ 64.68          & \ \ 81.86          & 23.30          & \ \ 50.32          & 107.30         & 132.10         & 12.47          & 25.77          & \ \ 48.08          & \ \ 58.29          & 15.27          & 29.27          & \ \ 71.54          & \ \ 96.65          \\
				Traj-GCN~\cite{mao2019learning}     & \ \ 8.97          & 19.87          & \ \ 43.35          & \textbf{\ \ 53.74} & 18.65          & \ \ 38.68          & \ \ 77.74          & \ \ 93.39          & 10.24          & 21.02          & \ \ 42.54          & \ \ 52.30          & 13.66          & 29.89          & \textbf{\ \ 66.62} & \textbf{\ \ 84.05} \\
				MSR-GCN               & \textbf{\ \ 8.61} & \textbf{19.65} & \textbf{\ \ 43.28} & \ \ 53.82          & \textbf{16.48} & \textbf{\ \ 36.95} & \textbf{\ \ 77.32} & \textbf{\ \ 93.38} & \textbf{10.10} & \textbf{20.74} & \textbf{\ \ 41.51} & \textbf{\ \ 51.26} & \textbf{12.79} & \textbf{29.38} & \ \ 66.95          & \ \ 85.01          \\ \hline
			\end{tabular}
			\begin{tabular}{c|cccc|cccc|cccc|cccc}
				\hline
				scenarios             & \multicolumn{4}{c|}{purchases}                                    & \multicolumn{4}{c|}{sitting}                                      & \multicolumn{4}{c|}{sittingdown}                                  & \multicolumn{4}{c}{takingphoto}                                 \\ \hline
				millisecond (ms)      & 80             & 160            & 320            & 400            & 80             & 160            & 320            & 400            & 80             & 160            & 320            & 400            & 80            & 160            & 320            & 400            \\ \hline
				Residual sup.~\cite{martinez2017human} & 36.33          & 60.30          & \ \ 86.53          & \ \ 95.92          & 42.55          & \ \ 81.40          & 134.70         & 151.78         & 47.28          & 85.95          & 145.75         & 168.86         & 26.10         & 47.61          & \ \ 81.40          & \ \ 94.73          \\
				DMGNN~\cite{li2020dynamic}        & 21.35          & 38.71          & \ \ 75.67          & \ \ 92.74          & 11.92          & \ \ 25.11          & \ \ 44.59          & \ \ 50.20          & 14.95          & 32.88          & \ \ 77.06          & \ \ 93.00          & 13.61         & 28.95          & \ \ 45.99          & \ \ 58.76          \\
				Traj-GCN~\cite{mao2019learning}     & 15.60          & 32.78          & \textbf{\ \ 65.72} & \textbf{\ \ 79.25} & 10.62          & \textbf{\ \ 21.90} & \ \ 46.33          & \ \ 57.91          & 16.14          & \textbf{31.12} & \textbf{\ \ 61.47} & \textbf{\ \ 75.46} & \textbf{\ \ 9.88} & \textbf{20.89} & \ \ 44.95          & \ \ 56.58          \\
				MSR-GCN               & \textbf{14.75} & \textbf{32.39} & \ \ 66.13          & \ \ 79.64          & \textbf{10.53} & \ \ 21.99          & \textbf{\ \ 46.26} & \textbf{\ \ 57.80} & \textbf{16.10} & 31.63          & \ \ 62.45          & \ \ 76.84          & \ \ 9.89          & 21.01          & \textbf{\ \ 44.56} & \textbf{\ \ 56.30} \\ \hline
			\end{tabular}
			\begin{tabular}{c|cccc|cccc|cccc|cccc}
				\hline
				scenarios             & \multicolumn{4}{c|}{waiting}                                      & \multicolumn{4}{c|}{walkingdog}                                   & \multicolumn{4}{c|}{walkingtogether}                              & \multicolumn{4}{c}{Average}                                      \\ \hline
				millisecond (ms)      & 80             & 160            & 320            & 400            & 80             & 160            & 320            & 400            & 80             & 160            & 320            & 400            & 80             & 160            & 320            & 400            \\ \hline
				Residual sup.~\cite{martinez2017human} & 30.62          & 57.82          & 106.22         & 121.45         & 64.18          & 102.10         & 141.07         & 164.35         & 26.79          & 50.07          & \ \ 80.16          & \ \ 92.23          & 34.66          & 61.97          & 101.08         & 115.49         \\
				DMGNN~\cite{li2020dynamic}        & 12.20          & 24.17          & \ \ 59.62          & \ \ 77.54          & 47.09          & \ \ 93.33          & 160.13         & 171.20         & 14.34          & 26.67          & \ \ 50.08          & \ \ 63.22          & 16.95          & 33.62          & \ \ 65.90          & \ \ 79.65          \\
				Traj-GCN~\cite{mao2019learning}     & 11.43          & 23.99          & \ \ 50.06          & \ \ 61.48          & 23.39          & \ \ 46.17          & \ \ 83.47          & \ \ 95.96          & 10.47          & 21.04          & \ \ 38.47          & \ \ 45.19          & 12.68          & 26.06          & \ \ 52.27          & \ \ 63.51          \\
				MSR-GCN               & \textbf{10.68} & \textbf{23.06} & \textbf{\ \ 48.25} & \textbf{\ \ 59.23} & \textbf{20.65} & \textbf{\ \ 42.88} & \textbf{\ \ 80.35} & \textbf{\ \ 93.31} & \textbf{10.56} & \textbf{20.92} & \textbf{\ \ 37.40} & \textbf{\ \ 43.85} & \textbf{12.11} & \textbf{25.56} & \textbf{\ \ 51.64} & \textbf{\ \ 62.93} \\ \hline
			\end{tabular}
		\end{center}
		\normalsize
		
		\vspace{-0.1cm}
	\end{table*}

	% human3.6m long-term
	\begin{table*}[]
		\vspace{-0.2cm}
		\begin{center}
			\renewcommand\tabcolsep{8.5pt}
			\scriptsize
			\caption{Comparisons for long-term prediction on 5 action categories of H3.6M and the averages. The best results are highlighted in bold.}
			\label{tab:h36m_long}
			\begin{tabular}{c|cc|cc|cc|cc|cc|cc}
				\hline
				scenarios        & \multicolumn{2}{c|}{walking}    & \multicolumn{2}{c|}{Eating}     & \multicolumn{2}{c|}{Smoking}    & \multicolumn{2}{c|}{Discussion}  & \multicolumn{2}{c|}{Directions}  & \multicolumn{2}{c}{average}     \\ \hline
				millisecond (ms) & 560            & 1000             & 560            & 1000             & 560            & 1000             & 560            & 1000              & 560            & 1000              & 560            & 1000             \\ \hline
				Residual sup.\cite{martinez2017human}    & 81.73          & 100.68         & 79.87          & 100.20         & 94.83          & 137.44         & 121.30         & 161.70          & 110.05         & 152.48          & 97.56          & 130.50         \\
				DMGNN \cite{li2020dynamic}            & 73.36          & 95.82          & 58.11          & 86.66          & 50.85          & 72.15          & \textbf{81.90} & 138.32          & 110.06         & 115.75          & 74.85          & 101.74         \\
				Traj-GCN \cite{mao2019learning}         & 54.05          & \textbf{59.75} & 53.39          & 77.75          & 50.74          & 72.62          & 91.61          & 121.53          & \textbf{71.01} & 101.79          & 64.16          & 86.69          \\
				MSR-GCN             & \textbf{52.72} & 63.04          & \textbf{52.54} & \textbf{77.11} & \textbf{49.45} & \textbf{71.64} & 88.59          & \textbf{117.59} & 71.18          & \textbf{100.59} & \textbf{62.89} & \textbf{86.00} \\ \hline
			\end{tabular}
		\end{center}
		\normalsize
	\end{table*}

	% mocap short-term
	\begin{table*}[]
		\vspace{-0.2cm}
		\begin{center}
			\renewcommand\tabcolsep{5pt}
			\scriptsize
			\caption{Comparisons for short-term prediction on 8 action categories of the CMU Mocap dataset. The best results are highlighted in bold.}
			\label{tab:mocap_short}
			\begin{tabular}{c|cccc|cccc|cccc|cccc}
				\hline
				scenarios             & \multicolumn{4}{c|}{basketball}                                   & \multicolumn{4}{c|}{basketball signal}                          & \multicolumn{4}{c|}{directing traffic}                           & \multicolumn{4}{c}{jumping}                                       \\ \hline
				millisecond (ms)      & 80             & 160            & 320            & 400            & 80            & 160           & 320            & 400            & 80            & 160            & 320            & 400            & 80             & 160            & 320            & 400            \\ \hline
				Residual sup.~\cite{martinez2017human} & 15.45          & 26.88          & 43.51          & \ \ 49.23          & 20.17         & 32.98         & 42.75          & 44.65          & 20.52         & 40.58          & \ \ 75.38          & \ \ 90.36          & 26.85          & 48.07          & 93.50          & 108.90         \\
				DMGNN~\cite{li2020dynamic}        & 15.57          & 28.72          & 59.01          & \ \ 73.05          & \ \ 5.03          & \ \ 9.28          & 20.21          & 26.23          & 10.21         & 20.90          & \ \ 41.55          & \ \ 52.28          & 31.97          & 54.32          & 96.66          & 119.92         \\
				Traj-GCN~\cite{mao2019learning}     & 11.68          & 21.26          & 40.99          & \ \ 50.78          & \ \ 3.33          & \ \ 6.25          & 13.58          & 17.98          & \ \ 6.92          & 13.69          & \ \ 30.30          & \ \ 39.97          & 17.18          & 32.37          & 60.12          & \ \ 72.55          \\
				MSR-GCN               & \textbf{10.28} & \textbf{18.94} & \textbf{37.68} & \textbf{\ \ 47.03} & \textbf{\ \ 3.03} & \textbf{\ \ 5.68} & \textbf{12.35} & \textbf{16.26} & \textbf{\ \ 5.92} & \textbf{12.09} & \textbf{\ \ 28.36} & \textbf{\ \ 38.04} & \textbf{14.99} & \textbf{28.66} & \textbf{55.86} & \textbf{\ \ 69.05} \\ \hline
			\end{tabular}
			\begin{tabular}{c|cccc|cccc|cccc|cccc}
				\hline
				scenarios             & \multicolumn{4}{c|}{running}                                      & \multicolumn{4}{c|}{soccer}                                       & \multicolumn{4}{c|}{walking}                                     & \multicolumn{4}{c}{washwindow}                                   \\ \hline
				millisecond (ms)      & 80             & 160            & 320            & 400            & 80             & 160            & 320            & 400            & 80            & 160            & 320            & 400            & 80            & 160            & 320            & 400            \\ \hline
				Residual sup.~\cite{martinez2017human} & 25.76          & 48.91          & 88.19          & 100.80         & 17.75          & 31.30          & 52.55          & 61.40          & 44.35         & 76.66          & 126.83         & 151.43         & 22.84         & 44.71          & 86.78          & 104.68         \\
				DMGNN~\cite{li2020dynamic}        & 17.42          & 26.82          & 38.27          &\ \  40.08          & 14.86          & 25.29          & 52.21          & 65.42          & \ \ 9.57          & 15.53          & \ \ 26.03          & \ \ 30.37          & \ \ 7.93          & 14.68          & 33.34          & \ \ 44.24          \\
				Traj-GCN~\cite{mao2019learning}     & 14.53          & 24.20          & 37.44          & \ \ 41.10          & 13.33          & 24.00          & 43.77          & 53.20          & \ \ 6.62          & 10.74          & \textbf{\ \ 17.40} & \textbf{\ \ 20.35} & \ \ 5.96          & 11.62          & \textbf{24.77} & \textbf{\ \ 31.63} \\
				MSR-GCN               & \textbf{12.84} & \textbf{20.42} & \textbf{30.58} & \textbf{\ \ 34.42} & \textbf{10.92} & \textbf{19.50} & \textbf{37.05} & \textbf{46.38} & \textbf{\ \ 6.31} & \textbf{10.30} & \ \ 17.64          & \ \ 21.12          & \textbf{\ \ 5.49} & \textbf{11.07} & 25.05          & \ \ 32.51          \\ \hline
			\end{tabular}
			
		\end{center}
		\normalsize
		\vspace{-0.1cm}
	\end{table*}

	% mocap long-term
	\begin{table}[]
		\begin{center}
			\vspace{-0.2cm}
			\renewcommand\tabcolsep{8pt}
			\scriptsize
			\caption{Comparisons for long-term prediction at 1000ms on 8 action categories of the CMU Mocap dataset.}
			\label{tab:mocap_long}
			\begin{tabular}{c|p{25pt}|p{25pt}|p{25.5pt}|p{28.5pt}}
				\hline
				scenarios     & basket & bas\_sig & dir\_tra & jumping         \\ \hline
				Residual sup.\cite{martinez2017human} & \textbf{\ \ 72.83} & \ \ 60.57          & 153.12          & 162.84          \\
				DMGNN \cite{li2020dynamic}         & 138.62         & \ \ 52.04          & 111.23          & 224.63          \\
				Traj-GCN \cite{mao2019learning}      & \ \ 97.99          & \ \ 54.00              & 114.16             & 127.41          \\
				MSR-GCN          & \ \ 86.96 & \textbf{\ \ 47.91}     & \textbf{111.04}    & \textbf{124.79} \\ \hline
			\end{tabular}
			
			\begin{tabular}{c|p{24pt}|p{25pt}|p{25pt}|p{23pt}}
				\hline
				scenarios     & \multicolumn{1}{l|}{running} & \multicolumn{1}{l|}{soccer} & \multicolumn{1}{l|}{walking} & \multicolumn{1}{l}{washwin} \\ \hline
				Residual sup.\cite{martinez2017human} & 158.19         & 107.37         & 194.33         & 202.73         \\
				DMGNN \cite{li2020dynamic}        & \textbf{\ \ 46.40} & 111.90         & \ \ 67.01          & \ \ 82.84          \\
				Traj-GCN \cite{mao2019learning}      & \ \ 51.73          & 108.26         & \textbf{\ \ 34.41} & \textbf{\ \ 66.95} \\
				MSR-GCN          & \ \ 48.03          & \textbf{\ \ 99.32} & \ \ 39.70          & \ \ 71.30          \\ \hline
			\end{tabular}
		\end{center}
		\normalsize
		\vspace{-0.1cm}
	\end{table}

	%------------------------------------------------------------------------
	\subsection{Datasets Setup}
	
	The \textbf{H3.6M} dataset~\cite{ionescu2013human3} consists of seven subjects S1, S5, S6, S7, S8, S9, and S11, and each one contains 15 action categories. We transform the original data from exponential mapping (expmap) format to the 3D joint coordinate space, downsample the original pose sequence by 2 along the time axis, and choose $22$ body joints from the original $32$ joints of a single pose. Like \cite{martinez2017human, li2020dynamic,mao2019learning}, we use the data of S5 and S11 as test and validation dataset respectively, and the rest data is used for training. We use four scales in descending and ascending section, which contains 22, 12, 7, and 4 joints respectively. 
	
	The \textbf{CMU Mocap} dataset is another commonly used dataset for human pose prediction, which includes 8 action categories. A single pose has 38 body joints in the original dataset, among which we choose 25 and abstract to 12, 7, and 4 joints. Other details are similar to H3.6M.

	\subsection{Comparison Settings}
	
	\textbf{Metrics.} Mean Per Joint Position Error (MPJPE) in millimeter is the most widely used evaluation metric. Supposing the predicted pose sequence is $\hat{X}_{1:T}$ and the corresponding ground truth is $X_{1:T}$, then the MPJPE loss is 
	\begin{equation}
		\mathcal{L}_\text{MPJPE} = \frac{1}{J \times T} \sum_{t=1}^{T} \sum_{j=1}^{J} {\left \| \hat{p}_{j, t} - p_{j, t} \right \|}^{2},
	\end{equation} 
	where $\hat{p}_{j, t} \in{\mathbb{R}^{3}}$ represents the predicted $j$-th joint position in frame $t$, and $p_{j, t}$ is the corresponding ground truth.
	
	\textbf{Baselines.} We compare our approach with three state-of-the-art baselines, {\em i.e.}, denoted as Residual sup.~\cite{martinez2017human}, DMGNN~\cite{li2020dynamic}, and Traj-GCN~\cite{mao2019learning}, respectively. The~\cite{martinez2017human} is based on RNN, and the rest two are based on GCNs. Specifically, ~\cite{li2020dynamic} builds a dynamic multi-scale graph convolution neural network, and~\cite{mao2019learning} transforms the original data from 3D coordinate space to frequency space.

	\textbf{Random test batch \vs full test set.} All the compared three works~\cite{martinez2017human,li2020dynamic,mao2019learning} evaluate their methods on just one randomly selected single batch data of size 8 for each action category. We argue that such little test data is not enough to accurately evaluate the performance of the compared approaches. This has also been questioned in \cite{pavllo2019modeling}. To alleviate this problem, we modify their published codes and retrain the networks to use the whole test dataset in 3D coordinate space to evaluate the MPJPE. Experimental results with the same evaluation manner from prior works can also be found in the supplemental material.
	
	\textbf{Unifying input and output length.} Methods of~\cite{martinez2017human,li2020dynamic} require 50 historical observed poses to predict 25 future poses, while~\cite{mao2019learning} predicts 25 future poses by just 10 poses. All the experiments in this paper follow the way of~\cite{mao2019learning}.
	
	\subsection{Results}
	
	To validate the prediction performance of MSR-GCN, we show the quantitative and qualitative results of MSR-GCN for 400ms short-term ({\em i.e.}, 10 frames) and 1000ms long-term ({\em i.e.}, 25 frames) predictions on H3.6M and CMU Mocap, and compare MSR-GCN with the state-of-the-art methods. 
	
\begin{figure}[t]
		\vspace{-0.5cm}
		\centering
		\includegraphics[width=0.9\linewidth]{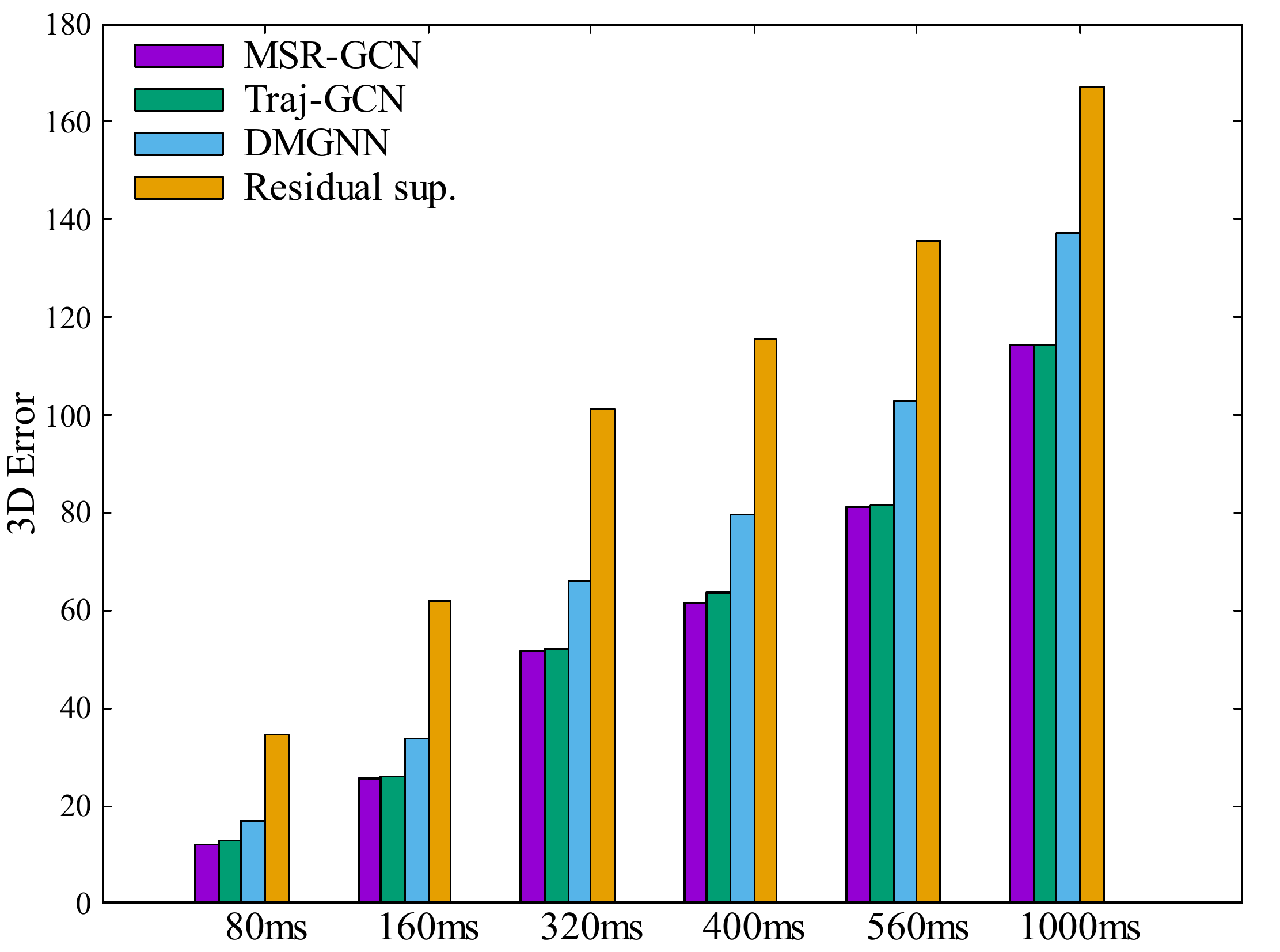}
		\caption{Comparison of average prediction error over all action categories at different forecast times on the H3.6M dataset.}
		\label{fig:other-hist}
		\vspace{-0.4cm}
	\end{figure}

\begin{figure}[t]
		\centering
		\includegraphics[width=1.0\linewidth]{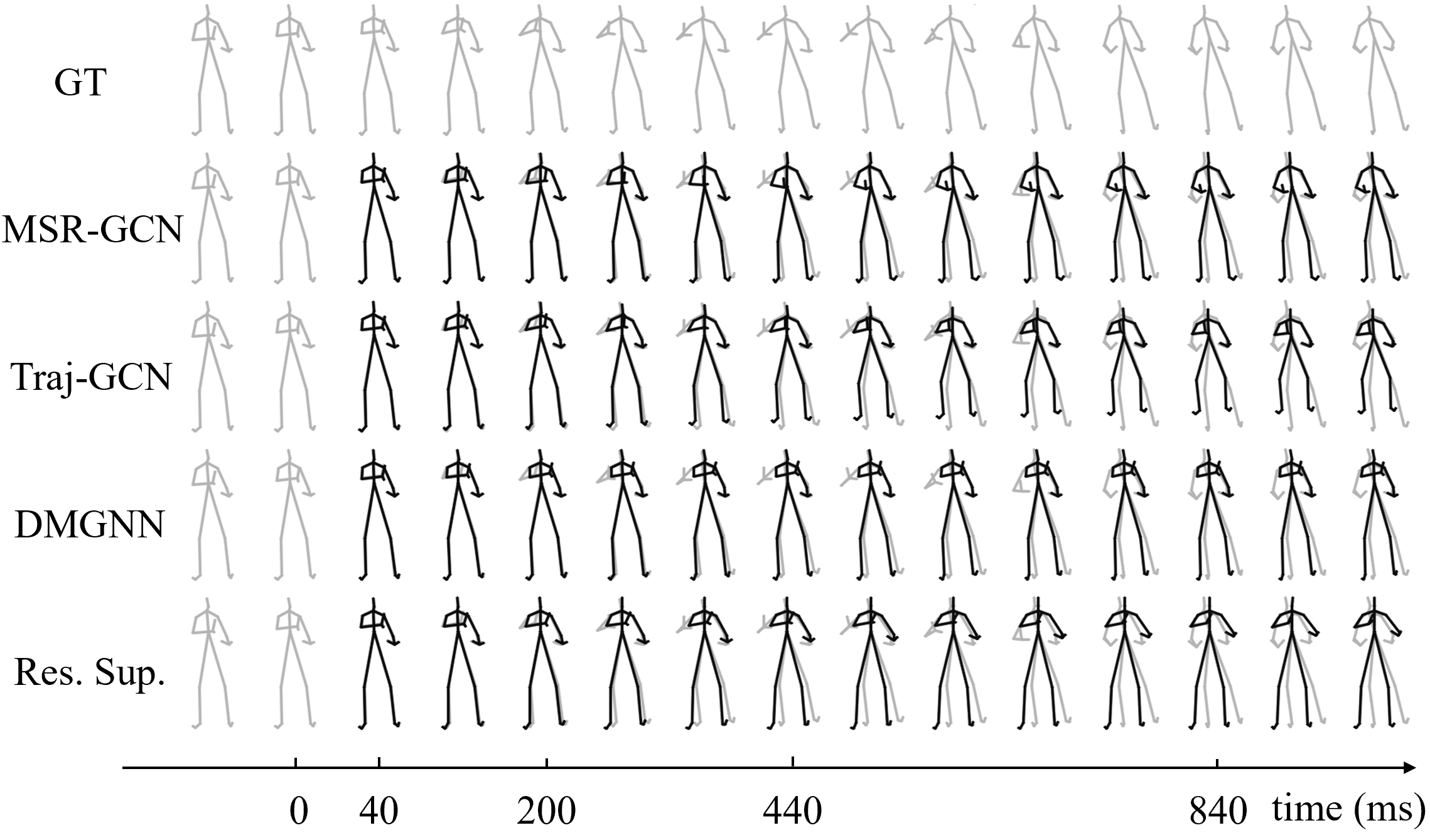}
		\caption{Visualization of predicted poses of different methods on a sample of the H3.6M dataset.}
		\label{fig:pose-other}
		\vspace{-0.3cm}
	\end{figure}

	\textbf{Results on H3.6M.} The quantitative comparisons for both short-term and long-term prediction results are presented in Table~\ref{tab:h36m_short1} and Table~\ref{tab:h36m_long} respectively. Apparently, the three GCN-based approaches are much better than the RNN-based method Residual sup.~\cite{martinez2017human}, which validates the effectiveness of GCNs for human motion prediction. Among the three GCN-based methods, Traj-GCN is better than DMGNN, while MSR-GCN is better than Traj-GCN, overall. For a more intuitive comparison, we plot the average prediction error over all kinds of actions at different forecast times in Figure~\ref{fig:other-hist}, which clearly shows that MSR-GCN outperforms the compared three methods. Figure~\ref{fig:pose-other} shows an example of the predicted poses for different methods. In this example, with the increase of the forecast time, the result of MSR-GCN becomes better than those of the others.
	
	\textbf{Results on CMU Mocap.} The same comparisons are conducted on the CMU Mocap dataset, as shown in Table \ref{tab:mocap_short} and Table \ref{tab:mocap_long}. MSR-GCN gets the best average performance at all short-term forecast times. For long-term prediction, \ie, predicting the frame up to 1000ms, MSR-GCN achieves the best results on four kinds of actions. For other actions, the prediction errors of our method are always the second best and are very close to the best ones.
	
	\textbf{Performance gains analysis and reasoning.} The above results show that MSR-GCN outperforms the compared methods. Here, we explain in detail the reasons and sources of performance gains.
	
	Firstly, during experiments, we find that inferring residuals between input and target poses is much easier than predicting the target poses. The average errors on the CMU dataset in Table~\ref{tab:GR} show that global residual (GR) leads to noticeable performance gains for both Traj-GCN and our method (MSR-GCN). Nevertheless, ours without GR still clearly outperforms other baselines without GR (Traj-GCN w/o residual and DMGNN), showing the significance of our model design.
	
	\begin{table}[]
		\vspace{-0.2cm}
		\begin{center}
			\scriptsize{
				\setlength{\tabcolsep}{1mm}{
					\caption{Effects of the global residual on the CMU Mocap dataset.}
					\label{tab:GR}
					\begin{tabular}{c|c|c|c|c}
						\hline
						DMGNN~\cite{li2020dynamic} & \makecell{Traj-GCN~\cite{mao2019learning}  w/o GR} & Traj-GCN~\cite{mao2019learning} & \makecell{Ours w/o GR} & Ours \\
						\hline
						53.05 & 49.82 & 39.75 & 46.92 & \textbf{37.28}\\
						\hline
			\end{tabular}}}
		\end{center}
		\vspace{-0.2cm}
	\end{table}
	
	Secondly, we compare our method with Traj-GCN, Traj-GCN w/o DCT, and a single-scale version of our method named MSR-GCN-1L on the CMU dataset. As shown in Table~\ref{tab:structure}, the performance gain led by DCT is 0.55, while that of our multi-scale strategy is 3.15, manifesting the effectiveness of our multi-scale architecture.
	
	\begin{table}[]	
		\vspace{-0.1cm}
		\begin{center}
			\scriptsize{
				\setlength{\tabcolsep}{2mm}{
					\caption{Comparison between the multi-scale architecture of MSR-GCN and the DCT components of Traj-GCN~\cite{mao2019learning} on CMU dataset.}
					\label{tab:structure}
					\begin{tabular}{c|c|c|c}
						\hline
						\makecell{Traj-GCN~\cite{mao2019learning} w/o DCT} & Traj-GCN~\cite{mao2019learning} & MSR-GCN-1L & MSR-GCN \\
						\hline
						40.30 & 39.75 & 40.43 & 37.28\\
						\hline
			\end{tabular}}}
		\end{center}
		\vspace{-0.2cm}
	\end{table}
	
\begin{figure}[]
		\begin{center}
			%H36M
			\scriptsize{
				\hspace{0.1cm}
				\includegraphics[height=0.35\columnwidth]{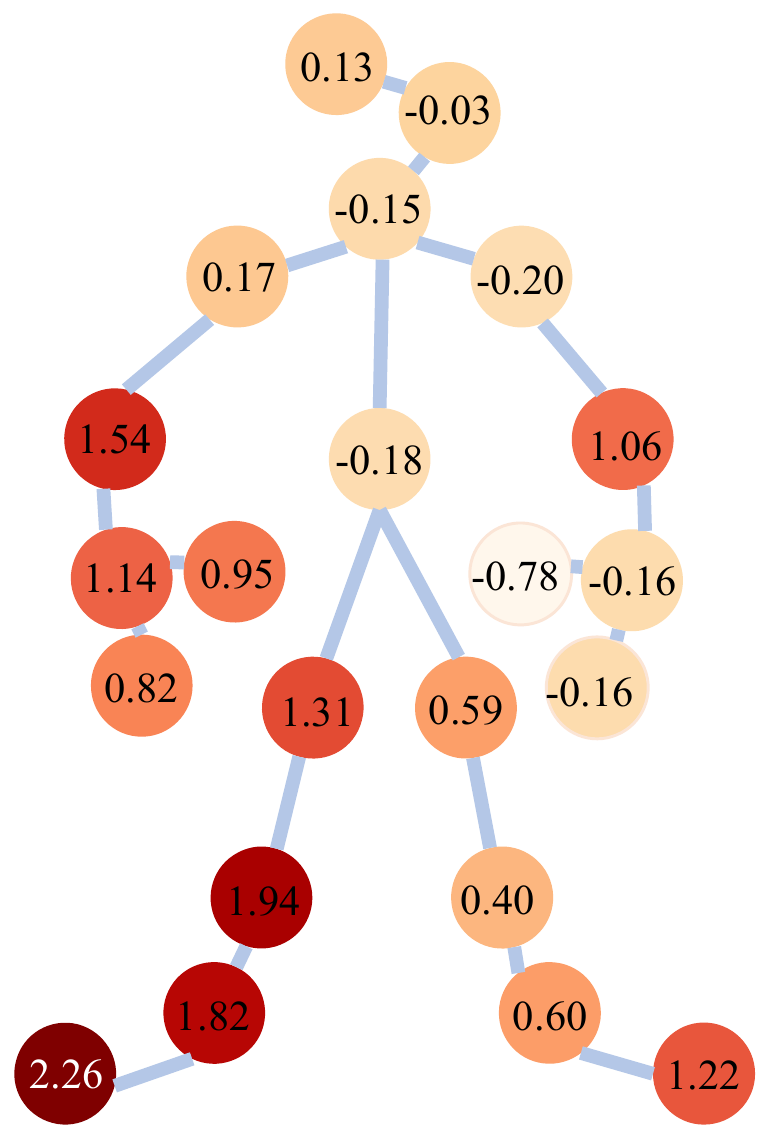}
				\includegraphics[height=0.35\columnwidth]{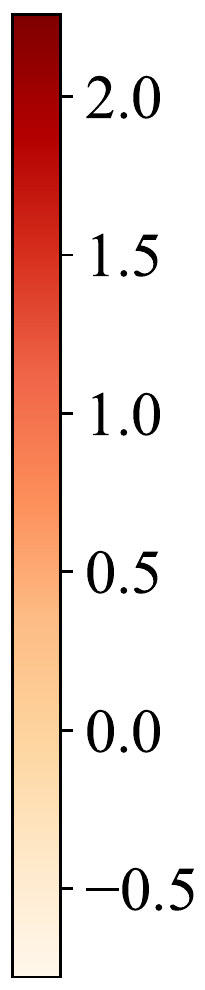}
				\hspace{0.4cm}
				%CMU
				\hspace{0.4cm}
				\includegraphics[height=0.35\columnwidth]{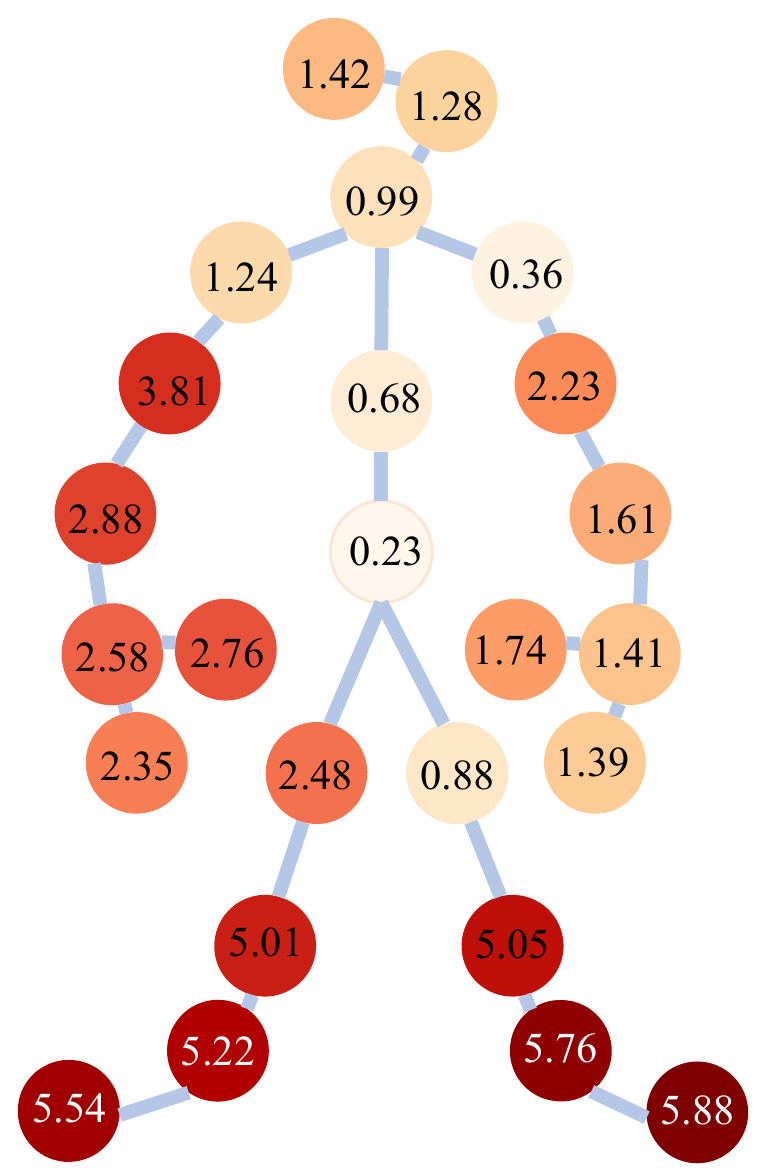}
				\includegraphics[height=0.35\columnwidth]{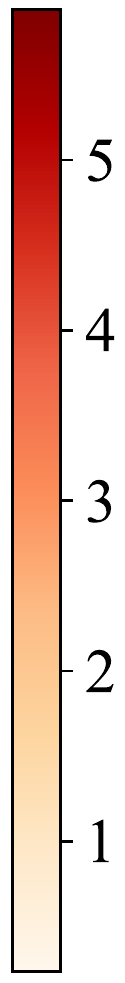}\\
			}
		\end{center}
		\vspace{-0.2cm}
		\caption{Average performance gain over Traj-GCN~\cite{mao2019learning} of joints on H36M (left) and CMU (right).}
		\label{fig:performance_gain_joint}
		\vspace{-0.5cm}
	\end{figure}

	Thirdly, we examine the performance gain of MSR-GCN over Traj-GCN for each joint, finding that larger performance gains are achieved for joints of limbs, as shown in Figure~\ref{fig:performance_gain_joint} where deeper red color means higher performance gain. Since joints on the limbs usually have higher motion frequency, the figure indicates that our method can better handle high-frequency motions.
	
	More analysis can be found in the supplemental material.
	
	% performance_gain_joint
	
	\subsection{Ablation Study}
	\begin{table*}[]
		\vspace{-0.5cm}
		\begin{center}
			\renewcommand\tabcolsep{4.5pt}
			\scriptsize
			\vspace{-0.3cm}
			\caption{Ablation studies on the number of scale levels, intermediate losses, residual GCNs \vs residual fully connected layers, and different grouping manners. Results are obtained on the CMU Mocap dataset. On average, all the designs of our model contribute to its accuracy.}
			\label{tab:ablation_table}
			\begin{tabular}{c|ccccccc|ccccc|ccccc}
				\hline
				\multicolumn{8}{c|}{}                                                                    & \multicolumn{5}{c|}{running}          & \multicolumn{5}{c}{soccer}     \\ \hline
				& s1 & s2 & s3 & \multicolumn{1}{c|}{s4} & \multicolumn{1}{c|}{inter-loss} & GCB & FCL & 80    & 160   & 320   & 400   & 1000  & 80   & 160  & 320   & 400   & 1000  \\ \hline
				MSR-GCN & \checkmark  & \checkmark  & \checkmark  & \multicolumn{1}{c|}{\checkmark}  & \multicolumn{1}{c|}{\checkmark}          & \checkmark   &       & \textbf{12.84}    & \textbf{20.42}    & \textbf{30.58}    & \textbf{34.42}    & \ \ \textbf{48.03}    & \textbf{10.92}    & \textbf{19.50}    & \textbf{37.05}    & \textbf{46.38}   & \textbf{99.32}    \\
				MSR-GCN w/o inter-loss&\checkmark  & \checkmark  & \checkmark  & \multicolumn{1}{c|}{\checkmark}  & \multicolumn{1}{c|}{}           & \checkmark   &        & 13.20    & 21.20    & 32.69    & 36.02    & \ \ 51.65    & 11.03    & 19.81    & 38.93    & 48.84   & 101.36   \\
				MSR-GCN-3L&\checkmark  & \checkmark  & \checkmark  & \multicolumn{1}{c|}{}   & \multicolumn{1}{c|}{\checkmark}          & \checkmark   &        & 13.60 & 22.79 & 35.87 & 39.58 & \ \ 49.60  & 11.02 & 19.84 & 38.49 & 48.26 & 107.17 \\
				MSR-GCN-2L&\checkmark  & \checkmark  &    & \multicolumn{1}{c|}{}   & \multicolumn{1}{c|}{\checkmark}          & \checkmark   &        & 14.30    & 23.37    & 38.95    & 45.11    & \ \ 73.26    & 10.93    & 19.62    & 38.44    & 48.30   & 106.35   \\
				MSR-GCN-1L&\checkmark  &    &    & \multicolumn{1}{c|}{}   & \multicolumn{1}{c|}{\checkmark}          & \checkmark   &       & 14.24    & 24.21    & 39.06    & 43.60    & \ \ 74.52    & 11.55    & 21.37    & 43.26    & 55.00   & 123.69 \\
				MSR-FCL&\checkmark  & \checkmark  & \checkmark  & \multicolumn{1}{c|}{\checkmark}  & \multicolumn{1}{c|}{\checkmark}          &     & \checkmark      & 13.33 & 24.29 & 43.58 & 50.01 & \ \ 61.90 & 12.16 & 22.83 & 46.49 & 59.04 & 132.47 \\ \hline
			\end{tabular}
			
			\begin{tabular}{c|ccccccc|ccccc|ccccc}
				\hline
				\multicolumn{8}{c|}{}                                                                    & \multicolumn{5}{c|}{walking}          & \multicolumn{5}{c}{jumping}     \\ \hline
				& s1 & s2 & s3 & \multicolumn{1}{c|}{s4} & \multicolumn{1}{c|}{inter-loss} & GCB & FCL & 80    & 160   & 320   & 400   & 1000  & 80   & 160  & 320   & 400   & 1000  \\ \hline
				MSR-GCN &\checkmark  & \checkmark  & \checkmark  & \multicolumn{1}{c|}{\checkmark}  & \multicolumn{1}{c|}{\checkmark}          & \checkmark   &       & \ \ \textbf{6.31}     & \textbf{10.30}    & 17.64    & 21.12    & \ \ 39.70  & 14.99 & 28.66 & \textbf{55.86} & \textbf{69.05} & \textbf{124.79} \\
				MSR-GCN w/o inter-loss&\checkmark  & \checkmark  & \checkmark  & \multicolumn{1}{c|}{\checkmark}  & \multicolumn{1}{c|}{}           & \checkmark   &        &  \ \ 6.36     & 10.33    & \textbf{17.05}    & \textbf{20.04}    & \ \ \textbf{34.67}  & \textbf{14.65} & \textbf{28.22} & 56.43 & 70.07 & 125.69 \\
				MSR-GCN-3L&\checkmark  & \checkmark  & \checkmark  & \multicolumn{1}{c|}{}   & \multicolumn{1}{c|}{\checkmark}          & \checkmark   &        &\ \ 6.62  & 10.91 & 18.10 & 21.19 & \ \ 42.72 & 14.98 & 28.89 & 57.69 & 71.60 & 128.62 \\
				MSR-GCN-2L &\checkmark  & \checkmark  &    & \multicolumn{1}{c|}{}   & \multicolumn{1}{c|}{\checkmark}          & \checkmark   &        & \ \ 7.87     & 13.41    & 23.16    & 27.63    & \ \ 52.31 & 15.21 & 29.67 & 59.85 & 74.31 & 128.10 \\
				MSR-GCN-1L&\checkmark  &    &    & \multicolumn{1}{c|}{}   & \multicolumn{1}{c|}{\checkmark}          & \checkmark   &        & \ \ 6.73     & 11.09    & 17.94    & 20.95    & \ \ 37.21 & 15.49 & 29.73 & 58.94 & 73.10 & 131.72 \\ 
				MSR-FCL & \checkmark  & \checkmark  & \checkmark  & \multicolumn{1}{c|}{\checkmark}  & \multicolumn{1}{c|}{\checkmark}          &     & \checkmark      & \ \ 7.19  & 12.58 & 23.15 & 28.00 & \ \ 52.77 & 15.14 & 29.89 & 61.31 & 76.49 & 139.01 \\ \hline
			\end{tabular}
			
		\end{center}
		\normalsize
		
		\vspace{-0.3cm}
	\end{table*}
	
	The influences of several key elements of our proposed model, such as the number of the scale levels, the intermediate supervision losses, the residual GCNs, and the multi-scale grouping manner, are investigated on the CMU Mocap dataset to provide a deeper understanding of our approach. Specifically, we modify MSR-GCN to obtain five ablation variants of it: (1) MSR-GCN w/o inter-loss: the MSR-GCN without intermediate supervision losses, (2) MSR-GCN-3L: the MSR-GCN with three pose scales (note that the original MSR-GCN has four scales), (3) and (4) MSR-GCN-2L, and MSR-GCN-1L with two scales and one scale respectively, (5) MSR-FCL: replace the residual GCNs by residual fully connected layers. 
	
	\textbf{Effects of multi-scale architecture.} To study the effectiveness of the multi-scale mechanism of the proposed architecture, we conduct experiments on the three-scale, two-scale and one-scale variants of MSR-GCN. The comparison results are shown in Table \ref{tab:ablation_table}. Please see the rows corresponding to MSR-GCN, MSR-GCN-3L, MSR-GCN-2L, and MSR-GCN-1L. In most cases, MSR-GCN is the best, followed by MSR-GCN-3L, MSR-GCN-2L, and MSR-GCN-1L. As an example, for the action of running, the prediction error of the four variants at time 320ms are 30.58, 35.87, 38.95, and 39.06, respectively. These experiments demonstrate the effectiveness of our multi-scale architecture.
	
	\textbf{Effects of intermediate supervisions.} The effects of intermediate losses are analyzed by removing the ``End GCNs'' of the second, the third, and the fourth scale from MSR-GCN. Please see the two rows corresponding MSR-GCN and MSR-GCN w/o inter-loss in Table~\ref{tab:ablation_table} to compare the two variants. In most cases, MSR-GCN is better than MSR-GCN w/o inter-loss, which demonstrates the necessity of the intermediate supervisions. Although some exceptions happen on ``walking'' and ``jumping'', the differences between the two variants are very small.
	
	\textbf{Effects of residual GCNs.} We replace all the residual GCNs with plain networks comprising residual fully connected layers (FCL) to analyze the effects of the residual GCNs. Please see the rows corresponding to MSR-GCN and MSR-FCL of Table~\ref{tab:ablation_table}. The experimental results show that MSR-GCN is better than MSR-FCL by a large margin. This strongly validates the importance of GCNs for high-quality pose prediction.

	\textbf{Effects of different multi-scale grouping manners. \label{sec:ablationgroupingmanner}}  In default, we group the human joints in the way shown in Figure~\ref{fig:fine-coarse} for skeletons of H3.6M. The default grouping manner for CMU can be found in the supplemental material. In Table~\ref{tab:grouping_manners}, we test the performance of our method with different grouping strategies on CMU, including 25-10-5-3 which means there are 25 joints for the finest-scale skeleton and 3 joints for the coarsest scale (please refer to the supplemental material for the manually specified joint groups), and three random groupings of the default 25-12-7-4. As shown, our default grouping produces better average results.
	
	\begin{table}[]
		\vspace{-0.2cm}
		\begin{center}
			\scriptsize{
				\setlength{\tabcolsep}{0.85mm}{
					\caption{Comparison of average errors of different grouping manners on the CMU dataset.}
					\label{tab:grouping_manners}
					\begin{tabular}{c|cccc|c}
						\hline
						\multirow{2}{*}{Grouping} & \multicolumn{4}{c|}{25-12-7-4}                                                                            & 25-10-5-3 \\ \cline{2-6} 
						& \multicolumn{1}{c|}{ Specified (default)} & \multicolumn{1}{c|}{Random 1} & \multicolumn{1}{c|}{Random 2} & Random 3 & Specified  \\ \hline
						Avg. Error $\downarrow$                        & \multicolumn{1}{c|}{\textbf{37.28}}    & \multicolumn{1}{c|}{41.15}    & \multicolumn{1}{c|}{45.77}    & 47.04    & 40.99     \\ \hline
			\end{tabular}}}
		\end{center}
		\vspace{-0.2cm}
	\end{table}

	\begin{figure}[t]
		\vspace{-0.2cm}
		\centering
		\includegraphics[width=1.0\linewidth]{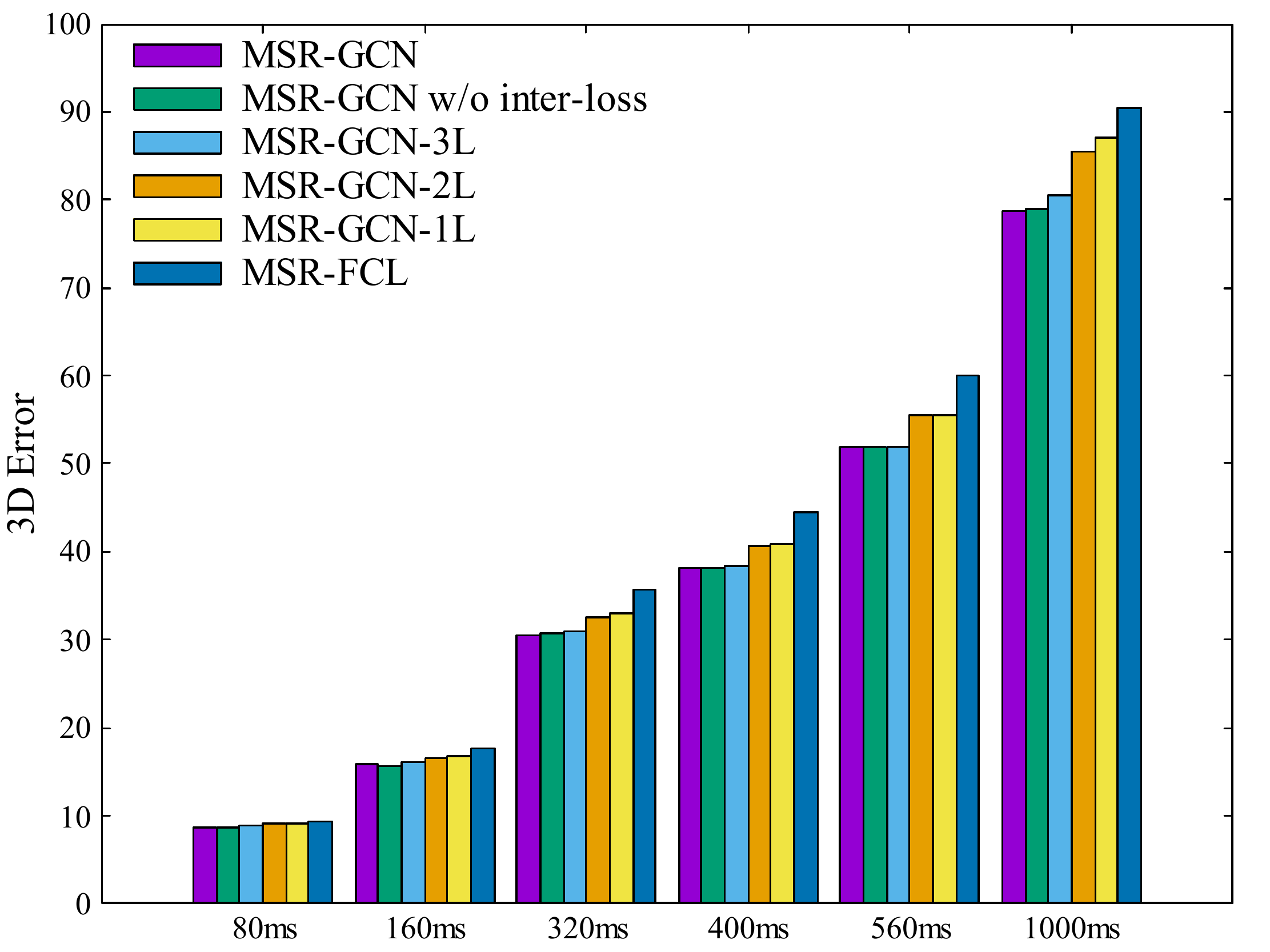}
		\caption{Comparison of average errors over all kinds of actions of different ablation variants at different forecast times on CMU.}
		\label{fig:ablation_final}
		\vspace{-0.4cm}
	\end{figure}
	
\begin{figure}[t]
		\vspace{-0.1cm}
		\centering
		\includegraphics[width=1.0\linewidth]{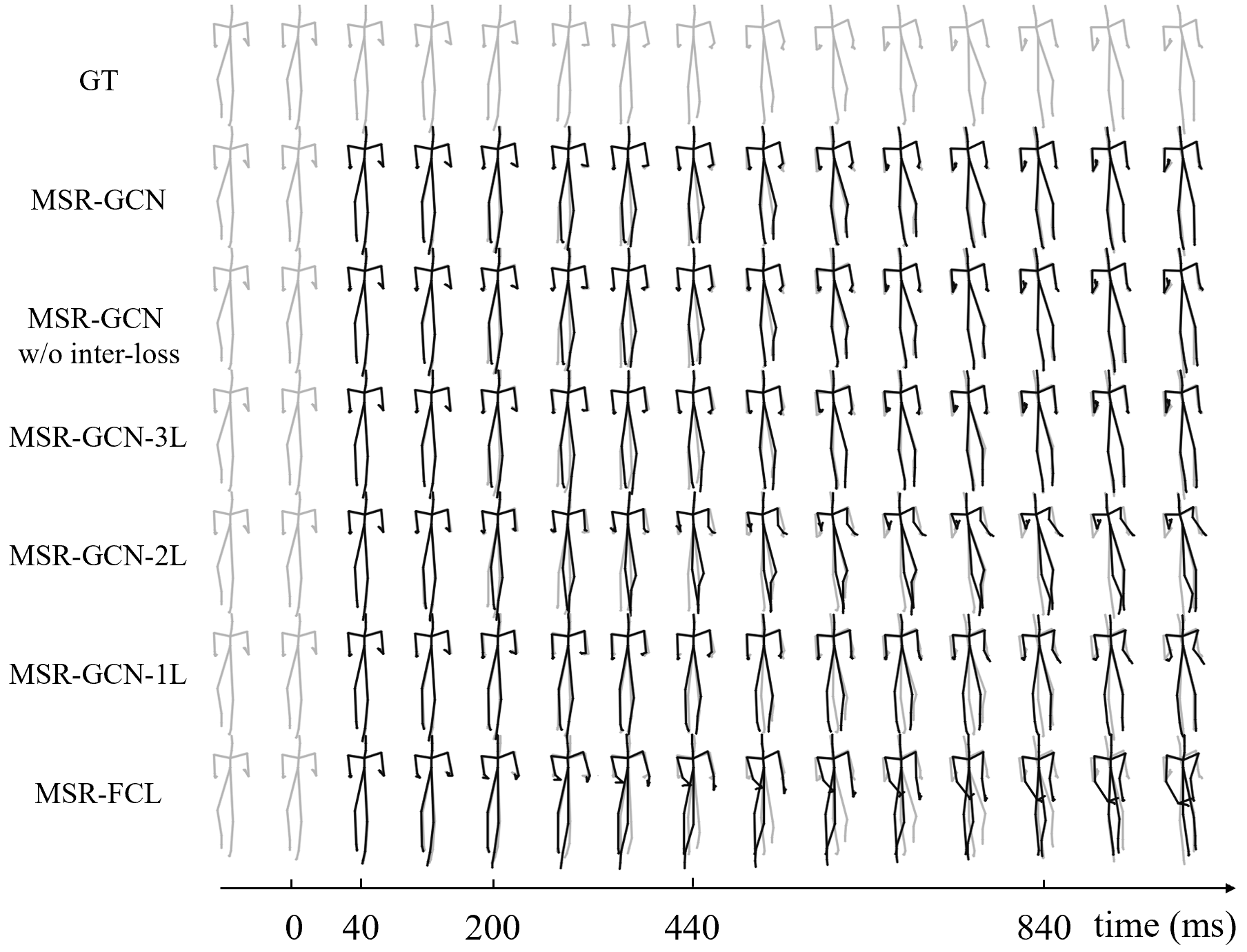}
		\caption{Visualization of predicted poses of different ablation variants on a sample of the CMU Mocap dataset.}
		\label{fig:pose-ablation}
		\vspace{-0.4cm}
	\end{figure}

	More visualizations are shown in Figure~\ref{fig:ablation_final} and Figure~\ref{fig:pose-ablation}. In Figure~\ref{fig:ablation_final}, we show the average prediction errors over all kinds of actions of different ablation variants at different forecast times on the CMU dataset. As can be seen, MSR-GCN is always better than its variants. In Figure~\ref{fig:pose-ablation}, we show an example of the predicted poses of different ablation variants, which clearly demonstrate that MSR-GCN is much better than MSR-GCN-2L, MSR-GCN-1L, and MSR-FCL, verifying the necessity of both the building blocks of GCNs and the multi-scale architecture.

	%-----------------------------------------------------------------------
	\section{Conclusion}
	
	In this paper, we build a multi-scale residual graph convolution network to effectively predict future human motion from observed histories. Losses are added to all the scales to provide intermediate supervision. We use a short observed historical pose sequence of 10 frames as input to predict 25 frames in the future. We test and compare the proposed method with previous state-of-the-art approaches on the whole test dataset. Our approach outperforms the state-of-the-art methods on two standard benchmark datasets. We will further explore the multi-scale grouping manners in the future.
	
	\section*{Acknowledgement}
	\vspace{-0.2cm}
	This research is sponsored in part by the National Natural Science Foundation of China (62072191, 61802453, 61972160), in part by the Natural Science Foundation of Guangdong Province (2019A1515010860, 2021A1515012301), and in part by the Fundamental Research Funds for the Central Universities (D2190670).
	
	%------------------------------------------------------------------------
	
	{\small
		\bibliographystyle{ieee_fullname}
		\bibliography{egbib_full}
	}
	
	\newpage
	
	\appendix
	\noindent{\LARGE{\textbf{Appendices}}}
	
	\section{Loss Function}
	
	We use $\ell_{2}$ loss to optimize MSR-GCN. Let the $j^{th}$ joint position in the $t^{th}$ frame at $s^{th}$ scale be $\hat{p}_{j, t}^{s}$, and the corresponding ground-truth be $p_{j, t}^{s}$, then the loss function for $N$ training pose sequences each having $J^{s}$ joints and $T$ frames is written as
	\begin{equation}
		\mathcal{L}^{s} = \frac{1}{N \times J^{s} \times T} \sum_{n=1}^{N} \sum_{j=1}^{J^{s}} \sum_{t=1}^{T} {\left \| \hat{p}_{j, t}^{s} - p_{j, t}^{s} \right \|}_{2}.
	\end{equation}
	The above loss is calculated at all $S$ scales and added up to optimize the proposed model, that is,
	\begin{equation}
		\mathcal{P}^{*} = \mathop{\arg\min_{\mathcal{P}}} \sum_{s=1}^{S} \lambda^{s} \mathcal{L}^{s},
	\end{equation}
	where $\mathcal{P}$ indicates network parameters, and $\lambda$ denotes hyper parameters and we set them as 1 for all scales.
	
	\section{Model Structure}
	\begin{table}[]
		\begin{center}
			\renewcommand\tabcolsep{3.5pt}
			% \footnotesize
			\scriptsize
			\caption{Detailed architecture of MSR-GCN.}
			\label{model}
			\begin{tabular}{c|c|c|c}
				\hline
				Module                          & Layers  & Output Size & Operations                                                                                     \\ \hline
				\makecell{\multirow{2}{*}{Start GCN}}      & GCL     & 66 $\times$ 64       & GCL,  A(66 $\times$ 66), W(35 $\times$ 64)   \\ \cline{2-4} 
				& GCN    & 66 $\times$ 64       & \begin{tabular}[c]{@{}c@{}}res-GCN with 2-layer GCLs \\ A(66 $\times$ 66), W(64 $\times$ 64)\end{tabular}    \\ \hline
				D0                              & GCNs    & 66 $\times$ 64       & \begin{tabular}[c]{@{}c@{}}3 $\times$ res-GCN each has 2-layer GCLs \\ A(66 $\times$ 66), W(64 $\times$ 64)\end{tabular}    \\ \hline
				\multirow{2}{*}{Downsampling 0} & Linear1 & 36 $\times$ 64       & linear transformation, W(66 $\times$ 36)                                                                \\ \cline{2-4} 
				& Linear2 & 36 $\times$ 128      & linear transformation,W(64 $\times$ 128)                                                                \\ \hline
				D1                              & GCNs    & 36 $\times$ 128      & \begin{tabular}[c]{@{}c@{}}3 $\times$ res-GCN each has 2-layer GCLs \\ A(36 $\times$ 36), W(128 $\times$ 128)\end{tabular}    \\ \hline
				\multirow{2}{*}{Downsampling 1} & Linear1 & 21 $\times$ 128      & linear transformation, W(36 $\times$ 21)                                                                \\ \cline{2-4} 
				& Linear2 & 21 $\times$ 256      & linear transformation,W(128 $\times$ 256)                                                               \\ \hline
				D2                              & GCNs    & 21 $\times$ 256      & \begin{tabular}[c]{@{}c@{}}3 $\times$ res-GCN each has 2-layer GCLs \\ A(21 $\times$ 21), W(256 $\times$ 256)\end{tabular}  \\ \hline
				\multirow{2}{*}{Downsampling 2} & Linear1 & 12 $\times$ 256      & linear transformation, W(21 $\times$ 12)                                                                \\ \cline{2-4} 
				& Linear2 & 12 $\times$ 512      & linear transformation,W(256 $\times$ 512)                                                               \\ \hline
				D3                              & GCNs    & 12 $\times$ 512      & \begin{tabular}[c]{@{}c@{}}3 $\times$ res-GCN each has 2-layer GCLs \\ A(12 $\times$ 12), W(512 $\times$ 512)\end{tabular}  \\ \hline
				A3                              & GCNs    & 12 $\times$ 512      & \begin{tabular}[c]{@{}c@{}}3 $\times$ res-GCN each has 2-layer GCLs \\ A(12 $\times$ 12), W(512 $\times$ 512)\end{tabular} \\ \hline
				\multirow{2}{*}{Upsampling 2}   & Linear1 & 21 $\times$ 512       & linear transformation, W(12 $\times$ 21)                                                                \\ \cline{2-4} 
				& Linear2 & 21 $\times$ 256      & linear transformation,W(512 $\times$ 256)                                                               \\ \hline
				A2                              & GCNs    & 21 $\times$ 256      & \begin{tabular}[c]{@{}c@{}}3 $\times$ res-GCN each has 2-layer GCLs \\ A(21 $\times$ 21), W(256 $\times$ 256)\end{tabular}  \\ \hline
				\multirow{2}{*}{Upsampling 1}   & Linear1 & 36 $\times$ 256       & linear transformation, W(21 $\times$ 36)                                                                \\ \cline{2-4} 
				& Linear2 & 36 $\times$ 128     & linear transformation,W(256 $\times$ 128)                                                               \\ \hline
				A1                              & GCNs    & 36 $\times$ 128      & \begin{tabular}[c]{@{}c@{}}3 $\times$ res-GCN each has 2-layer GCLs \\ A(36 $\times$ 36), W(128 $\times$ 128)\end{tabular}  \\ \hline
				\multirow{2}{*}{Upsampling 0}   & Linear1 & 66 $\times$ 128      & linear transformation, W(36 $\times$ 66)                                                                \\ \cline{2-4} 
				& Linear2 & 66 $\times$ 64       & linear transformation,W(128 $\times$ 64)                                                                \\ \hline
				A0                              & GCNs    & 66 $\times$ 64       & \begin{tabular}[c]{@{}c@{}}3 $\times$ res-GCN each has 2-layer GCLs \\ A(66 $\times$ 66), W(64 $\times$ 64)\end{tabular}    \\ \hline
				\multirow{2}{*}{E0}             & GCN    & 66 $\times$ 64       & \begin{tabular}[c]{@{}c@{}}res-GCN with 2-layer GCLs \\ A(66 $\times$ 66), W(64 $\times$ 64)\end{tabular}    \\ \cline{2-4} 
				& GCL     & 66 $\times$ 35       & GCL,  A(66 $\times$ 66), W(64 $\times$ 35)                                                                       \\ \hline
				\multirow{2}{*}{E1}             & GCN    & 36 $\times$ 128      & \begin{tabular}[c]{@{}c@{}}res-GCN with 2-layer GCLs \\ A(36 $\times$ 36), W(128 $\times$ 128)\end{tabular}  \\ \cline{2-4} 
				& GCL     & 36 $\times$ 35       & GCL,  A(36 $\times$ 36),   W(128 $\times$ 35)                                                                    \\ \hline
				\multirow{2}{*}{E2}             & GCN    & 21 $\times$ 256      & \begin{tabular}[c]{@{}c@{}}res-GCN with 2-layer GCLs \\ A(21 $\times$ 21), W(256 $\times$ 256)\end{tabular}  \\ \cline{2-4} 
				& GCL     & 21 $\times$ 35       & GCL,  A(21 $\times$ 21),   W(256 $\times$ 35)                                                                    \\ \hline
				\multirow{2}{*}{E3}             & GCN    & 12 $\times$ 512      & \begin{tabular}[c]{@{}c@{}}res-GCN with 2-layer GCLs \\ A(12 $\times$ 12), W(512 $\times$ 512)\end{tabular}   \\ \cline{2-4} 
				& GCL     & 12 $\times$ 35       & GCL,  A(12 $\times$ 12),   W(512 $\times$ 35)                                                                    \\ \hline
			\end{tabular}
		\end{center}
		\normalsize
		\vspace{-0.5cm}
	\end{table}
	
	The detailed MSR-GCN model structure is shown in Table~\ref{model}.  As mentioned in the paper, our proposed approach is composed of three kinds of GCNs, called ``Start GCNs'', ``Descending (D0-D3)/Ascending (A0-A3) GCNs'', and ``End GCNs (E0-E3)''. 
	
	The most basic building block is the Graph Convolution Layer (GCL), which consists of a graph convolution layer, a batch normalization layer, a tanh activation layer, and a dropout layer (with rate 0.1). A graph convolution layer has an adjacency matrix $A$ and parameters $W$.

	Each GCN is composed of 2 GCLs. The size of $A$ and $W$ of these GCLs are shown in the table. We use linear layers for downsampling and upsampling. The sizes of the parameters in these linear layers are also shown in the table. In the third column of the table, we give the output size of the corresponding layer. Please refer to the source code at \href{https://github.com/Droliven/MSRGCN}{https://github.com/Droliven/MSRGCN} for more information.

	%------------------------------------------------------------------------
	
	\section{Different Multi-Scale Grouping Manners}
	
\begin{figure}[]
		\centering
		\includegraphics[width=0.9\linewidth]{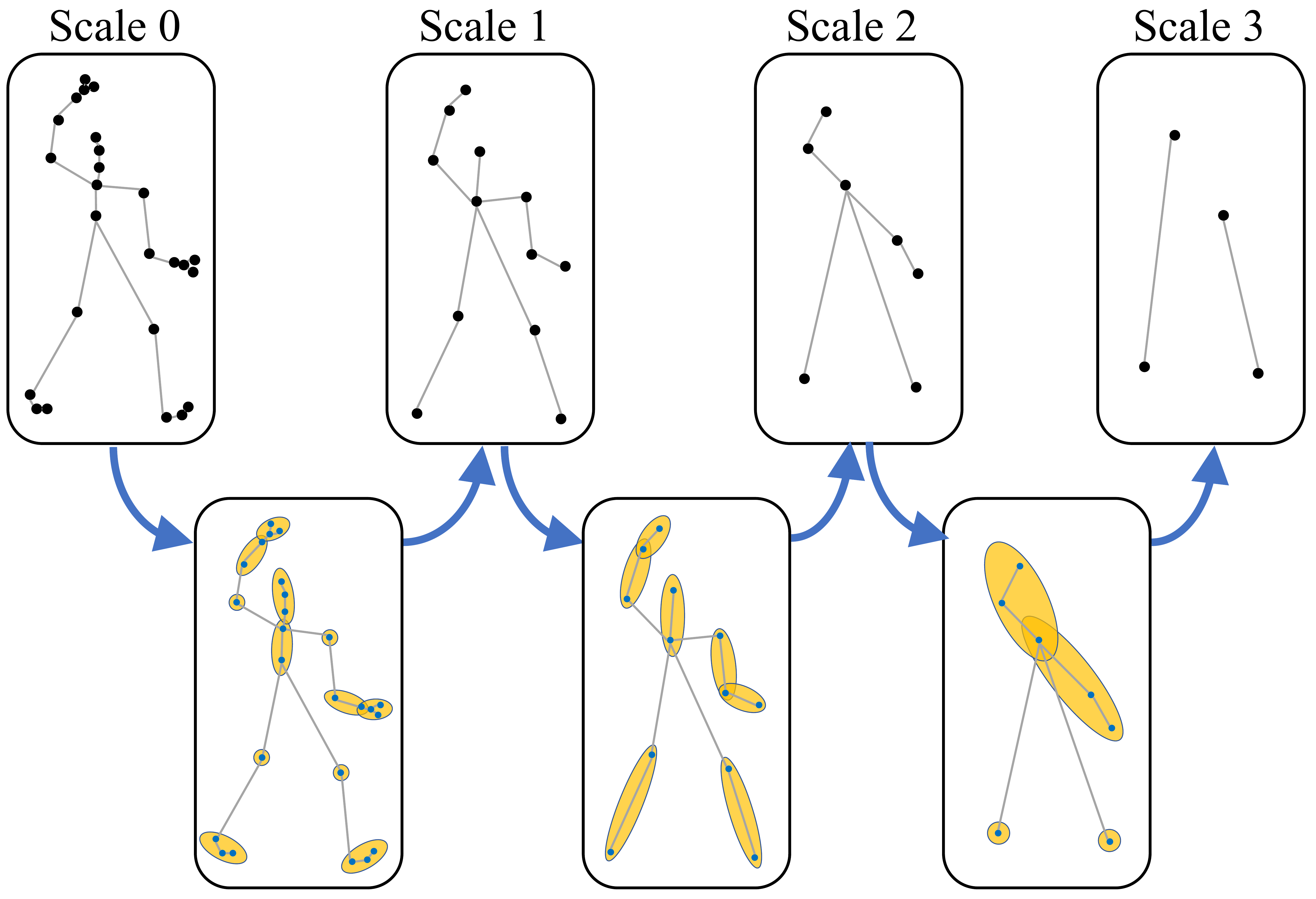}
		\caption{The default grouping manner of 25-12-7-4 for the CMU Mocap dataset.}
		\label{fig:fine_coarse_251274}
	\end{figure}

\begin{figure}[]
		\centering
		\includegraphics[width=0.9\linewidth]{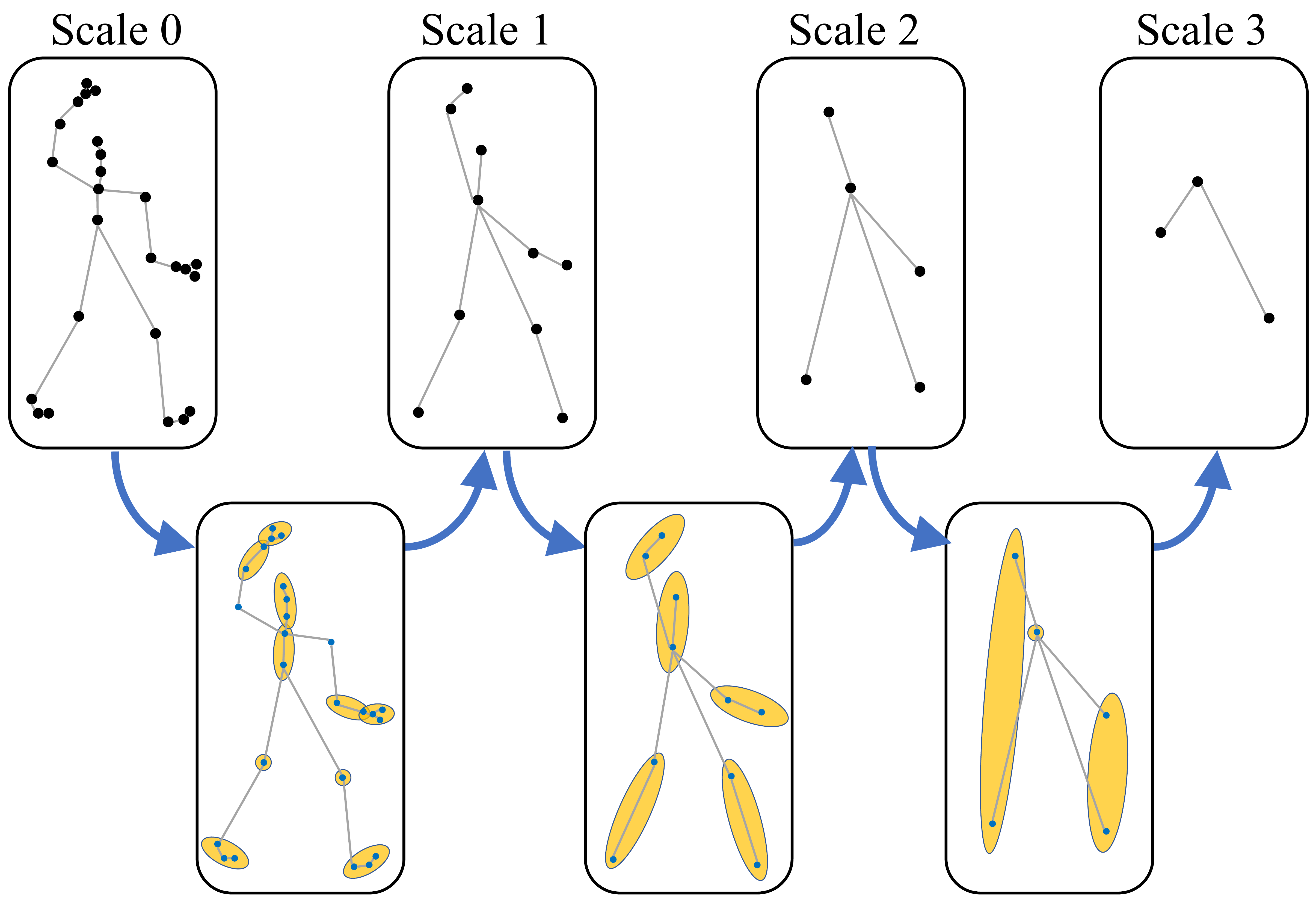}
		\caption{The manually specified 25-10-5-3 grouping manner for the CMU Mocap dataset.}
		\label{fig:fine_coarse_251053}
	\end{figure}

	%In default, we group the human joints in the way shown in Figure 1 in the main paper for skeletons of H3.6M. 
	The default grouping manner for CMU can be found in Figure~\ref{fig:fine_coarse_251274} in which there are 25 joints at the finest level and 12, 7, 4 joints in the subsequent coarser levels. We also trained MSR-GCN on CMU with other grouping manners, including three random grouping manners under the 25-12-7-4 manner, and the ``manually specified 25-10-5-3'' manner as shown in Figure~\ref{fig:fine_coarse_251053}. The performance of MSR-GCN under different grouping manners can be found in the paper. 
	
	%------------------------------------------------------------------------
	
	\section{More Results}
	
	\textbf{Comparison with Traj-GCN using error bar.} We have trained our method and Traj-GCN [33] five times with random seeds in order to compare their performance more thoroughly. As shown in Table~\ref{tab:error_bar}, the average prediction errors of our method are 58.37±0.43 and 37.52±0.48 on the datasets of Human3.6M and CMU. In comparison, [33] reports higher predictor errors and larger variances than our method, which are 59.93±0.91 on the Human3.6M and 40.56±0.50 on the CMU dataset respectively.
	
	\begin{table}[]
		\begin{center}
			\scriptsize{
				\setlength{\tabcolsep}{7mm}{
					\caption{Comparison of average prediction error with Traj-GCN [33] using error bar}
					\label{tab:error_bar}
					\begin{tabular}{c|c|c}
						\hline
						& H3.6M & CMU \\ \hline
						Traj-GCN [33] & 59.93 $\pm$ 0.91 & 40.56 $\pm$ 0.51 \\
						Ours     & \textbf{58.37$\pm$ 0.43}   & \textbf{37.52 $\pm$ 0.48} \\ \hline
			\end{tabular}}}
		\end{center}
		\normalsize
		\vspace{-0.3cm}
	\end{table}
	
	\textbf{Comparison with Traj-GCN at different forecast times.} We also compared MSR-GCN and Traj-GCN at different forecast times. As verified in Table~\ref{tab:performance_gain_timestep}, our method performs better than Traj-GCN in handling challenging long-term motion prediction.

	% performance_gain_timestep
	\begin{table}[]
		\begin{center}
			\scriptsize{
				\caption{Comparison with Traj-GCN at different forecast times.}
				\label{tab:performance_gain_timestep}
				\setlength{\tabcolsep}{2.8mm}{
					\begin{tabular}{c|cccccc}
						\hline
						Time (ms) & 80 & 160 & 320 & 40 & 560 & 1000 \\
						\hline
						Human3.6M & 0.56 & 0.51 & 0.64 & 0.58 &0.46 & 0.09 \\
						CMU & 1.22 & 2.19 & 3.98 & 2.84 & 2.40 & 3.23 \\
						\hline
			\end{tabular}}}
		\end{center}
		\normalsize
		\vspace{-0.3cm}
	\end{table}

	\textbf{Comparison using the evaluation method of [33].} In [33], the performance is evaluated on randomly selected 8 samples per action. The average prediction errors using the same evaluation method as [33] are shown in Table~\ref{tab:same_evaluation}. As can be seen, MSR-GCN also outperforms Traj-GCN.
	
	% same evaluation
	\begin{table}[]
		\begin{center}
			\scriptsize{
				\setlength{\tabcolsep}{2.5mm}{
					\caption{Average prediction errors using the evaluation method of [33].}
					\label{tab:same_evaluation}
					\begin{tabular}{c|cc|cc}
						\hline
						& \multicolumn{2}{c|}{H3.6M}                       & \multicolumn{2}{c}{CMU}                         \\ \cline{2-5} 
						& \multicolumn{1}{c|}{short-term} & long-term      & \multicolumn{1}{c|}{short-term} & long-term      \\ \hline
						Traj-GCN [33] & 37.35                           & 59.02          & 29.13                           & 45.06          \\
						Ours     & \textbf{36.36}                  & \textbf{57.84} & \textbf{24.81}                  & \textbf{40.81} \\ \hline
			\end{tabular}}}
		\end{center}
		\normalsize
		\vspace{-0.3cm}
	\end{table}
	
\end{document}